\documentclass[10pt,twocolumn,letterpaper]{article}

\usepackage{amsmath}
\usepackage{amssymb}
\usepackage{style/cvpr}
\usepackage{epsfig}
\usepackage{gensymb}
\usepackage{graphicx}
\usepackage{multirow}
\usepackage{booktabs}

\graphicspath{{./images/}}

\usepackage{mathtools}
\usepackage{subcaption}
\usepackage{times}
\usepackage[hyphens]{url}

\usepackage{style/ownstyles}
\definecolor{neural_renderer}{rgb}{0.4, 0.6, 0.4}
\definecolor{classifier}{rgb}{0.58, 0.43, 0.38}
\definecolor{red_dashed_line}{rgb}{0.75, 0.0, 0.0}
\definecolor{correct}{rgb}{0.07, 0.54, 0.31}
\definecolor{incorrect}{rgb}{0.72, 0.1, 0.08}

\newcommand{\subsec}[1]{\noindent{\textbf{#1.}}}
\newcommand{\class}[1]{{\small\texttt{#1}}}
\newif\ifcomments%

\commentstrue%

\ifcomments%
\newcommand{\comments}[1]{#1}
\else
\newcommand{\comments}[1]{}
\fi

\DeclareMathOperator*{\argmax}{arg\,max}

\newcommand{\x}{\mathbf{x}}
\newcommand{\w}{\mathbf{w}}

\newcommand{\LL}{\mathcal{L}}

\newcommand{\R}{\mathbb{R}}

\newcommand{\bright}{\ensuremath{\mathsf{bright}}\xspace}
\newcommand{\medium}{\ensuremath{\mathsf{medium}}\xspace}
\newcommand{\dark}{\ensuremath{\mathsf{dark}}\xspace}

\newcommand{\layer}[1]{\ensuremath{\mathsf{#1}\xspace}}

\usepackage[pagebackref=true,breaklinks=true,letterpaper=true,colorlinks,bookmarks=false]{hyperref}

\cvprfinalcopy 

\def\cvprPaperID{2951} 

\ifcvprfinal\fi
\begin{document}

\newcommand{\papertitle}{Strike (with) a Pose: Neural Networks Are Easily Fooled \\by Strange Poses of Familiar Objects}
\title{\papertitle}

\author{Michael A. Alcorn\\
{\tt\small alcorma@auburn.edu}
\and
Qi Li\\
{\tt\small qzl0019@auburn.edu}
\and
Zhitao Gong\\
{\tt\small gong@auburn.edu}
\and
Chengfei Wang\\
{\tt\small czw0078@auburn.edu}
\and
Long Mai\\
{\tt\small malong@adobe.com}
\and
Wei-Shinn Ku\\
{\tt\small weishinn@auburn.edu}
\and
Anh Nguyen\\
{\tt\small anhnguyen@auburn.edu}
\and
{Auburn University}\hspace{1cm}{Adobe Inc.~~~~~~~}
}

\maketitle

\begin{abstract}

Despite excellent performance on stationary test sets, deep neural networks (DNNs) can fail to generalize to out-of-distribution (OoD) inputs, including natural, non-adversarial ones, which are common in real-world settings.
In this paper, we present a framework for discovering DNN failures that harnesses 3D renderers and 3D models.
That is, we estimate the parameters of a 3D renderer that cause a target DNN to misbehave in response to the rendered image.
Using our framework and a self-assembled dataset of 3D objects, we investigate the vulnerability of DNNs to OoD poses of well-known objects in ImageNet.
For objects that are readily recognized by DNNs in their canonical poses, DNNs incorrectly classify 97\% of their pose space.
In addition, DNNs are highly sensitive to slight pose perturbations.
Importantly, adversarial poses transfer across models and datasets.
We find that 99.9\% and 99.4\% of the poses misclassified by Inception-v3 also transfer to the AlexNet and ResNet-50 image classifiers trained on the same ImageNet dataset, respectively, and 75.5\% transfer to the YOLOv3 object detector trained on MS COCO.
\end{abstract}

\section{Introduction}

For real-world technologies, such as self-driving cars~\cite{chen2015deepdriving}, autonomous drones~\cite{gandhi2017learning}, and search-and-rescue robots~\cite{sampedro2018fully}, the test distribution may be non-stationary, and new observations will often be out-of-distribution (OoD), \ie, not 
from the training distribution \cite{sugiyama2017dataset}.
However, machine learning (ML) models frequently assign wrong labels with high confidence to OoD examples, such as adversarial examples~\cite{szegedy2013intriguing,nguyen2015deep}---inputs specially crafted by an adversary to cause a target model to misbehave.
But ML models are also vulnerable to \emph{natural} OoD examples~\cite{lambert2016understanding,uber2017killed,tian2017deeptest,tesla2016killed}.
For example, when a Tesla autopilot car failed to recognize a white truck against a bright-lit sky---an unusual view that might be OoD---it crashed into the truck, killing the driver~\cite{tesla2016killed}.

\begin{figure}[t]
	\centering
	{
		(a)
		\hspace{1.5cm} (b)
		\hspace{1.5cm} (c)
		\hspace{1.7cm} (d)
	}
	\includegraphics[width=1.0\columnwidth]{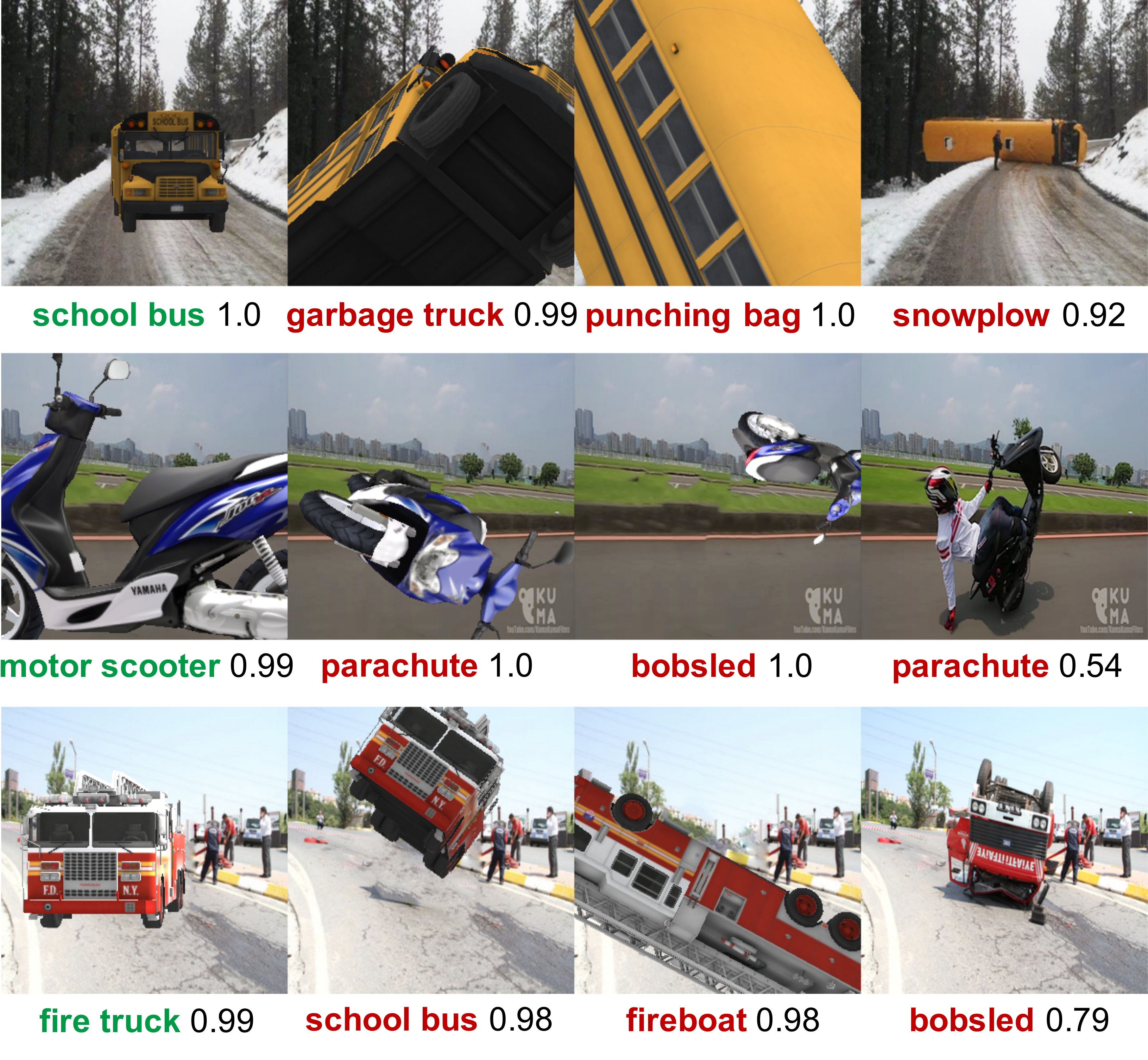}
	
	\caption{
		The Google Inception-v3 classifier~\cite{szegedy2016rethinking} correctly labels the canonical poses of objects (a), but fails to recognize out-of-distribution images of objects in unusual poses (b--d), including real photographs retrieved from the Internet (d).
		The left $3 \times 3$ images (a--c) are found by our framework and rendered via a 3D renderer.
		Below each image are its top-1 predicted label and confidence score.
	}\label{fig:teaser}
\end{figure}

Previous research has successfully used 3D graphics as a diagnostic tool for computer vision systems \cite{borji2016ilab, ozuysal2009pose, taylor2007ovvv, pinto2009far, qiu2017unrealcv}.
To understand natural Type II classification errors in DNNs, we searched for misclassified 6D poses (\ie, 3D translations and 3D rotations) of 3D objects.
Our results reveal that state-of-the-art image classifiers and object detectors trained on large-scale image datasets \cite{russakovsky2015imagenet,lin2014microsoft} misclassify most poses for many familiar training-set objects.
For example, DNNs predict the front view of a school bus---an object in the ImageNet dataset \cite{russakovsky2015imagenet}---extremely well (Fig.~\ref{fig:teaser}a) but fail to recognize the same object when it is too close or flipped over, \ie, in poses that are OoD yet exist in the real world (Fig.~\ref{fig:teaser}d).
However, a self-driving car needs to correctly estimate at least some attributes of an incoming, unknown object (instead of simply rejecting it~\cite{hendrycks2016baseline,scheirer2013toward}) to handle the situation gracefully and minimize damage.
Because road environments are highly variable~\cite{tesla2016killed,uber2017killed}, addressing this type of OoD error is a non-trivial challenge.

\begin{figure*}[t]
    \centering
    \includegraphics[width=0.90\linewidth]{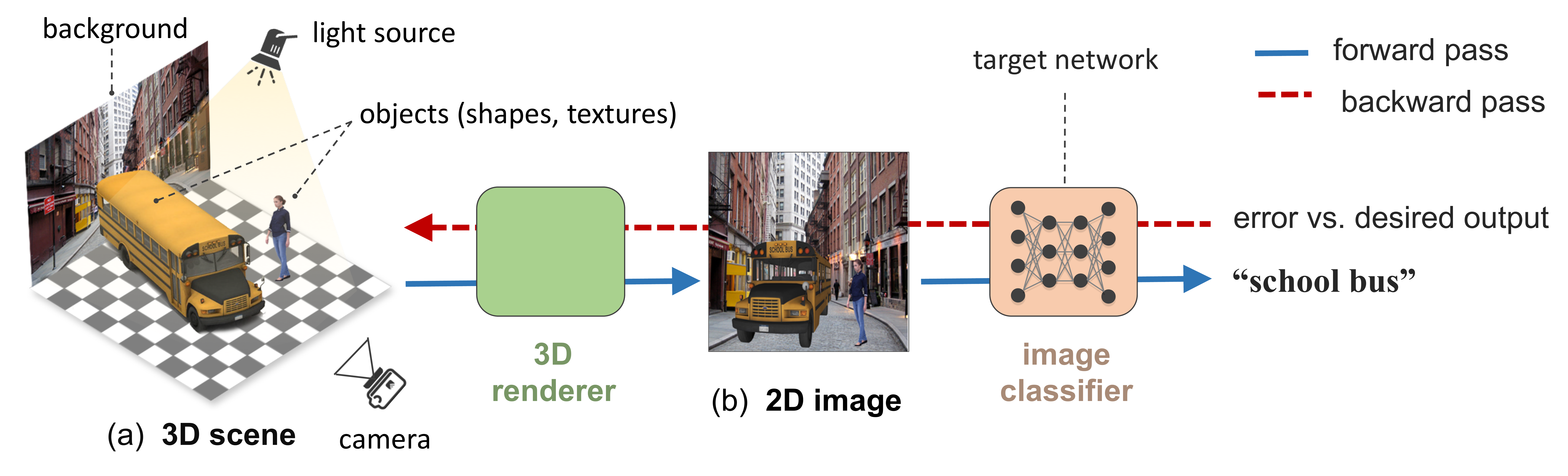}
    \caption{To test a target DNN, we build a 3D scene (a) that consists of 3D objects (here, a school bus and a pedestrian), lighting, a background scene, and camera parameters.
    Our 3D renderer renders the scene into a 2D image, which the image classifier labels \class{school bus}.
    We can estimate the pose changes of the school bus that cause the classifier to misclassify by (1) approximating gradients via finite differences; or (2) backpropagating (\textcolor{red_dashed_line}{red} dashed line) through a differentiable renderer.
    }\label{fig:concept}
\end{figure*}

In this paper, we propose a framework for finding OoD errors in computer vision models in which iterative optimization in the parameter space of a 3D renderer is used to estimate changes (\eg, in object geometry and appearance, lighting, background, or camera settings) that cause a target DNN to misbehave (Fig.~\ref{fig:concept}).
With our framework, we generated unrestricted 6D poses of 3D objects and studied how DNNs respond to 3D translations and 3D rotations of objects.
For our study, we built a dataset of 3D objects corresponding to 30 ImageNet classes relevant to the self-driving car application.
The code for our framework is available at 
\url{https://github.com/airalcorn2/strike-with-a-pose}.
In addition, we built a simple GUI tool that allows users to generate their own adversarial renders of an object.
Our main findings are:

\begin{itemize}
    \item ImageNet classifiers only correctly label $3.09\%$ of the entire 6D pose space of a 3D object, and misclassify many generated adversarial examples (AXs) that are human-recognizable (Fig.~\ref{fig:teaser}b--c).
    A misclassification can be found via a change as small as $10.31\degree$, $8.02\degree$, and $9.17\degree$ to the yaw, pitch, and roll, respectively.
    \item 99.9\% and 99.4\% of AXs generated against Inception-v3 transfer to the AlexNet and ResNet-50 image classifiers, respectively, and 75.5\% transfer to the YOLOv3 object detector.

    \item Training on adversarial poses generated by the 30 objects (in addition to the original ImageNet data) did not help DNNs generalize well to held-out objects in the same class.
\end{itemize}

In sum, our work shows that state-of-the-art DNNs perform \emph{image classification} well but are still far from true \emph{object recognition}.
While it might be possible to improve DNN robustness through adversarial training with many more 3D objects, we hypothesize that future ML models capable of visual reasoning may instead benefit from better incorporation of 3D information.

\section{Framework}

\subsection{Problem formulation}

Let $f$ be an image classifier  that maps an image $\x \in \R^{H\times W\times C}$ onto a softmax probability distribution over 1,000 output classes~\cite{szegedy2016rethinking}.
%
%
Let $R$ be a 3D renderer that takes as input a set of parameters $\phi$ and outputs a render, \ie, a 2D image $R(\phi) \in \R^{H\times W\times C}$ (see Fig.~\ref{fig:concept}).
Typically, $\phi$ is factored into mesh vertices $V$, texture images $T$, a background image $B$, camera parameters $C$, and lighting parameters $L$, \ie, $\phi = \{V, T, B, C, L\}$~\cite{kato2018neural}.
To change the 6D pose of a given 3D object, we apply a 3D rotation and 3D translation, parameterized by $\w \in \R^6$, to the original vertices $V$ yielding a new set of vertices $V^*$.

Here, we wish to estimate only the pose transformation parameters $\w$ (while keeping all parameters in $\phi$ fixed) such that the rendered image $R(\w; \phi)$ causes the classifier $f$ to assign the highest probability (among all outputs) to an incorrect target output at index $t$.
Formally, we attempt to solve the below optimization problem:

\begin{equation}
\label{eq:min}
\w^* = \argmax_{\w}(f_t(R(\w; \phi)))
\end{equation}

\noindent
In practice, we minimize the cross-entropy loss $\LL$ for the target class.
Eq.~\ref{eq:min} may be solved efficiently via backpropagation if both $f$ and $R$ are differentiable, \ie, we are able to compute $\partial \LL/\partial\w$.
However, standard 3D renderers, \eg, OpenGL~\cite{woo1999opengl}, typically include many non-differentiable operations and cannot be inverted~\cite{marschner2015fundamentals}.
Therefore, we attempted two approaches: (1) harnessing a recently proposed differentiable renderer and performing gradient descent using its analytical gradients; and (2) harnessing a non-differentiable renderer and approximating the gradient via finite differences.

We will next describe the target classifier (Sec.~\ref{sec:target_dnn}), the renderers (Sec.~\ref{sec:renderers}), and our dataset of 3D objects (Sec.~\ref{sec:3d_object_dataset}) before discussing the optimization methods (Sec.~\ref{sec:methods}).



%

\subsection{Classification networks}
\label{sec:target_dnn}

We chose the well-known, pre-trained Google Inception-v3 \cite{Szegedy2015} DNN from the PyTorch model zoo \cite{torch2018vision} as the main image classifier for our study (the default DNN if not otherwise stated).
The DNN has a 77.45\% top-1 accuracy on the ImageNet ILSVRC 2012 dataset~\cite{russakovsky2015imagenet} of 1.2 million images corresponding to 1,000 categories.


\subsection{3D renderers}
\label{sec:renderers}

\subsec{Non-differentiable renderer}
We chose ModernGL~\cite{modernGL}
as our non-differentiable renderer.
ModernGL is a simple Python interface for the widely used OpenGL graphics engine.
ModernGL supports fast, GPU-accelerated rendering.

\subsec{Differentiable renderer}
To enable backpropagation through the non-differentiable rasterization process, Kato et al.~\cite{kato2018neural} replaced the discrete pixel color sampling step with a linear interpolation sampling scheme that admits non-zero gradients.
While the approximation enables gradients to flow from the output image back to the renderer parameters $\phi$, the render quality is lower than that of our non-differentiable renderer (see Fig.~\ref{fig:compare_tessellation} for a 
comparison).
Hereafter, we refer to the two renderers as NR and DR.


\subsection{3D object dataset}
\label{sec:3d_object_dataset}

\subsec{Construction}
Our main dataset consists of 30 unique 3D object models (purchased from many 3D model marketplaces) corresponding to 30 ImageNet classes relevant to a traffic environment (Fig.~\ref{fig:dataset_A}).
The 30 classes include 20 vehicles (\eg, \class{school bus} and \class{cab}) and 10 street-related items (\eg, \class{traffic light}). See Sec.~\ref{sec:SI_3d_object_dataset} for more details.

Each 3D object is represented as a mesh, \ie, a list of triangular faces, each defined by three vertices \cite{marschner2015fundamentals}.
The 30 meshes have on average 9,908 triangles (Table~\ref{tab:num_triangles}).
To maximize the realism of the rendered images, we used only 3D models that have high-quality 2D image textures.
We did not choose 3D models from public datasets, \eg, ObjectNet3D \cite{xiang2016objectnet3d}, because most of them do not have high-quality image textures.
That is, the renders of such models may be correctly classified by DNNs but still have poor realism.

\subsec{Evaluation}
We recognize that a reality gap will often exist between a render and a real photo.
Therefore, we rigorously evaluated our renders to make sure the reality gap was acceptable for our study.
From $\sim$100 initially-purchased 3D models, we selected the 30 highest-quality models using the evaluation method below.

First, we quantitatively evaluated DNN predictions on the renders.
For each object, we sampled 36 unique views (common in ImageNet) evenly divided into three sets.
For each set, we set the object at the origin, the up direction to $(0,1,0)$, and the camera position to $(0,0,-z)$ where $z = \{4, 6, 8\}$.
We sampled 12 views per set by starting the object at a $10^\circ{}$ yaw and generating a render at every $30^\circ{}$ yaw-rotation.
Across all objects and all renders, the Inception-v3 top-1 accuracy was $83.23\%$ (compared to $77.45\%$ on ImageNet images \cite{szegedy2016rethinking}) with a mean top-1 confidence score of $0.78$ (Table~\ref{tab:avg_accuracy_30obj}).
See Sec.~\ref{sec:SI_3d_object_dataset} for more details.

Second, we qualitatively evaluated the renders by comparing them to real photos.
We produced 116 (real photo, render) pairs via three steps: (1) we retrieved real photos of an object (\eg, a car) from the Internet; (2) we replaced the object with matching background content in Adobe Photoshop; and (3) we manually rendered the 3D object on the background such that its pose closely matched that in the reference photo.
Fig.~\ref{fig:dataset_B} shows example (real photo, render) pairs.
While discrepancies can be spotted in our side-by-side comparisons, we found that most of the renders passed our human visual Turing test if presented alone.

%

\subsection{Background images}

Previous studies have shown that image classifiers may be able to correctly label an image when foreground objects are removed (\ie, based on only the background content) \cite{zhu2016object}.
Because the purpose of our study was to understand how DNNs recognize an object itself, a non-empty background would have hindered our interpretation of the results.
Therefore, we rendered all images against a plain background with RGB values of $(0.485, 0.456, 0.406)$, \ie, the mean pixel of ImageNet images.
Note that the presence of a non-empty background should not alter our main qualitative findings in this paper---adversarial poses can be easily found against real background photos (Fig.~\ref{fig:teaser}).



\section{Methods}
\label{sec:methods}


We will describe the common pose transformations  (Sec.~\ref{sec:transformations}) used in the main experiments.
We were able to experiment with non-gradient methods because: (1) the pose transformation space $\R^6$ that we optimize in is fairly low-dimensional; and (2) although the NR is non-differentiable, its rendering speed is several orders of magnitude faster than that of DR.
In addition, our preliminary results showed that the objective function considered in Eq.~\ref{eq:min} is highly non-convex (see Fig.~\ref{fig:landscape}), therefore, it is interesting to compare (1) random search vs. (2) gradient descent using finite-difference (FD) approximated gradients vs. (3) gradient descent using the DR gradients.

\subsection{Pose transformations}
\label{sec:transformations}

We used standard computer graphics transformation matrices to change the pose of 3D objects \cite{marschner2015fundamentals}.
Specifically, to rotate an object with geometry defined by a set of vertices $V = \{v_i\}$, we applied the linear transformations in Eq.~\ref{eq:rot} to each vertex $v_{i} \in \R^3$:

\begin{equation}
\label{eq:rot}
v_{i}^{R} = R_{y}R_{p}R_{r}v_{i}
\end{equation}

\noindent
where $R_{y}$, $R_{p}$, and $R_{r}$ are the $3\times 3$ rotation matrices for yaw, pitch, and roll, respectively (the matrices can be found in Sec.~\ref{sec:trans-mat}).
We then translated the rotated object by adding a vector $T = \begin{bmatrix} x_{\delta} & y_{\delta} & z_{\delta} \end{bmatrix}^\top$ to each vertex:

\begin{equation}
v_{i}^{R,T} = T + v_{i}^{R}
\end{equation}

In all experiments, the center $c \in \R^3$ of the object was constrained to be inside a sub-volume of the camera viewing frustum.
That is, the $x$-, $y$-, and $z$-coordinates of $c$ were within $[-s,s]$, $[-s,s]$, and $[-28,0]$, respectively, with $s$ being the maximum value that would keep $c$ within the camera frame.
Specifically, $s$ is defined as:

\begin{equation} \label{eq:max_trans}
s = d \cdot\tan(\theta_{v})
\end{equation}

\noindent
where $\theta_{v}$ is one half the camera's angle of view (\ie, $8.213\degree$ in our experiments) and $d$ is the absolute value of the difference between the camera's $z$-coordinate and $z_{\delta}$.


\subsection{Random search}
\label{sec:random_search}

In reinforcement learning problems, random search (RS) can be surprisingly effective compared to more sophisticated methods \cite{such2017deep}.
For our RS procedure, instead of iteratively following some approximated gradient to solve the optimization problem in Eq.~\ref{eq:min}, we simply randomly selected a new pose in each iteration.
The rotation angles for the matrices in Eq.~\ref{eq:rot} were uniformly sampled from $(0, 2\pi)$.
$x_{\delta}$, $y_{\delta}$, and $z_{\delta}$ were also uniformly sampled from the ranges defined in Sec.~\ref{sec:transformations}.

\subsection{$z_{\delta}$-constrained random search}

Our preliminary RS results suggest the value of $z_{\delta}$ (which is a proxy for the object's size in the rendered image) has a large influence on a DNN's predictions.
Based on this observation, we used a $z_{\delta}$-constrained random search (ZRS) procedure both as an initializer for our gradient-based methods and as a naive performance baseline (for comparisons in Sec.~\ref{sec:comparing_methods}).
The ZRS procedure consisted of generating 10 random samples of $(x_{\delta}, y_{\delta}, \theta_{y}, \theta_{p}, \theta_{r})$ at each of 30 evenly spaced $z_{\delta}$ from $-28$ to $0$.

When using ZRS for initialization, the parameter set with the maximum target probability was selected as the starting point.
When using the procedure as an attack method, we first gathered the maximum target probabilities for each $z_{\delta}$, and then selected the best two $z_{\delta}$ to serve as the new range for RS.

\subsection{Gradient descent with finite-difference}
\label{sec:fd}

We calculated the first-order derivatives via finite central differences and performed vanilla gradient descent to iteratively minimize the cross-entropy loss $\LL$ for a target class.
That is, for each parameter $\w_{i}$, the partial derivative is approximated by:

\begin{equation} \label{eq:fd}
\frac{\partial \LL}{\partial \w_{i}} = \frac{\LL(\w_{i} + \frac{h}{2}) - \LL(\w_{i} - \frac{h}{2})}{h}
\end{equation}

\noindent
Although we used an $h$ of 0.001 for all parameters, a different step size can be used per parameter.
Because radians have a circular topology (\ie, a rotation of 0 radians is the same as a rotation of $2\pi$ radians, $4\pi$ radians, etc.), we parameterized each rotation angle $\theta_i$ as $(\cos(\theta_i), \sin(\theta_i))$---a technique commonly used for pose estimation~\cite{Osadchy2005} and inverse kinematics~\cite{Choi1992}---which maps the Cartesian plane to angles via the $atan2$ function.
Therefore, we optimized in a space of $3 + 2 \times 3 = 9$ parameters.

The approximate gradient $\nabla \LL$ obtained from Equation~\eqref{eq:fd} served as the gradient in our gradient descent.
We used the vanilla gradient descent update rule:

\begin{equation}
\w \coloneqq \w - \gamma{\nabla \LL}(\w)
\end{equation}

\noindent
with a learning rate $\gamma$ of 0.001 for all parameters and optimized for $100$ steps (no other stopping criteria).
\section{Experiments and results}

\subsection{Neural networks are easily confused by object rotations and translations}
\label{sec:easily_confused}


\begin{figure}[h]
\begin{subfigure}{\linewidth}
    \centering
    \includegraphics[width=\linewidth]{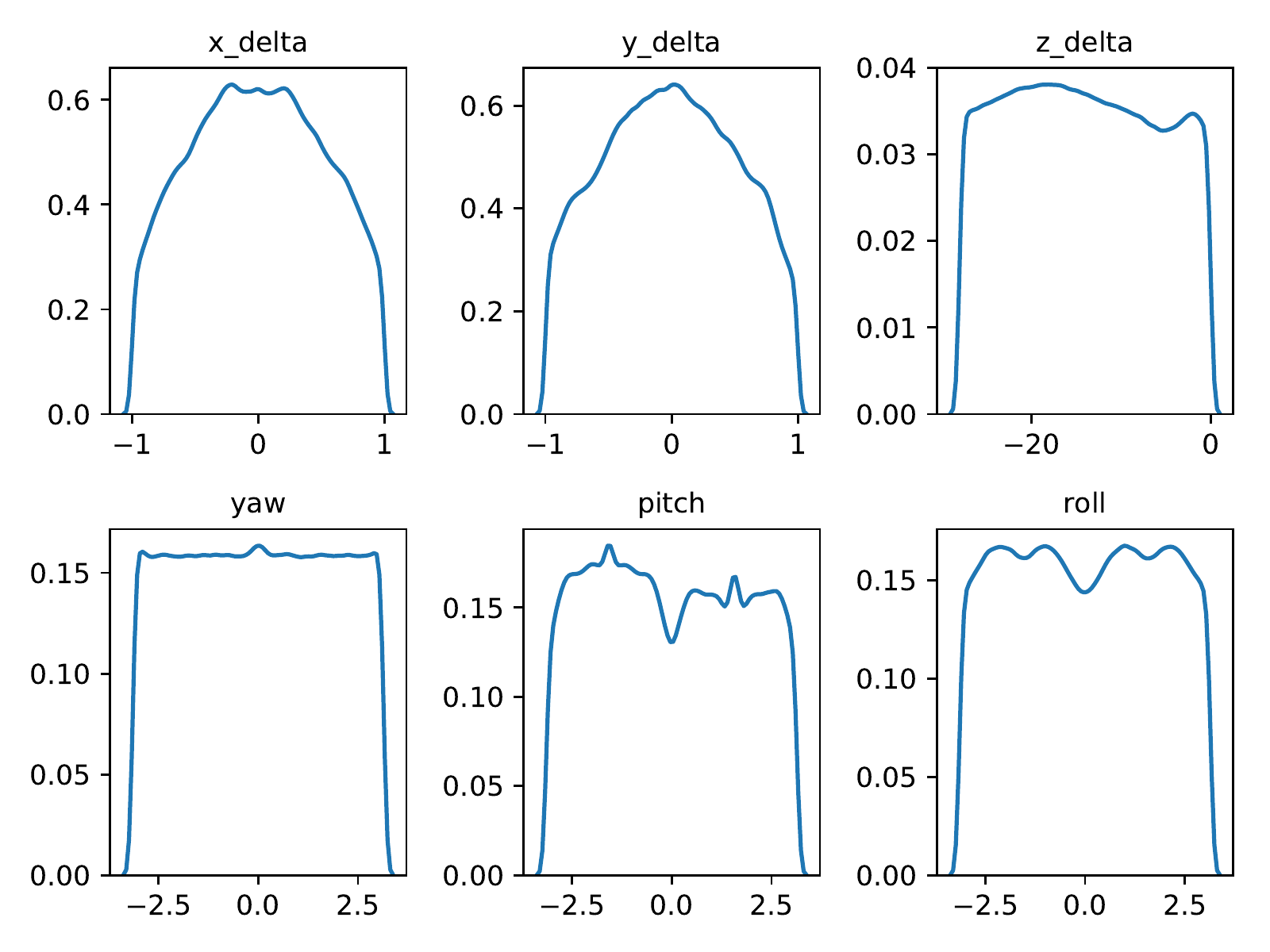}
    \caption{Incorrect classifications}\label{fig:high_conf_params}
\end{subfigure}
\begin{subfigure}{\linewidth}
    \centering
    \includegraphics[width=\linewidth]{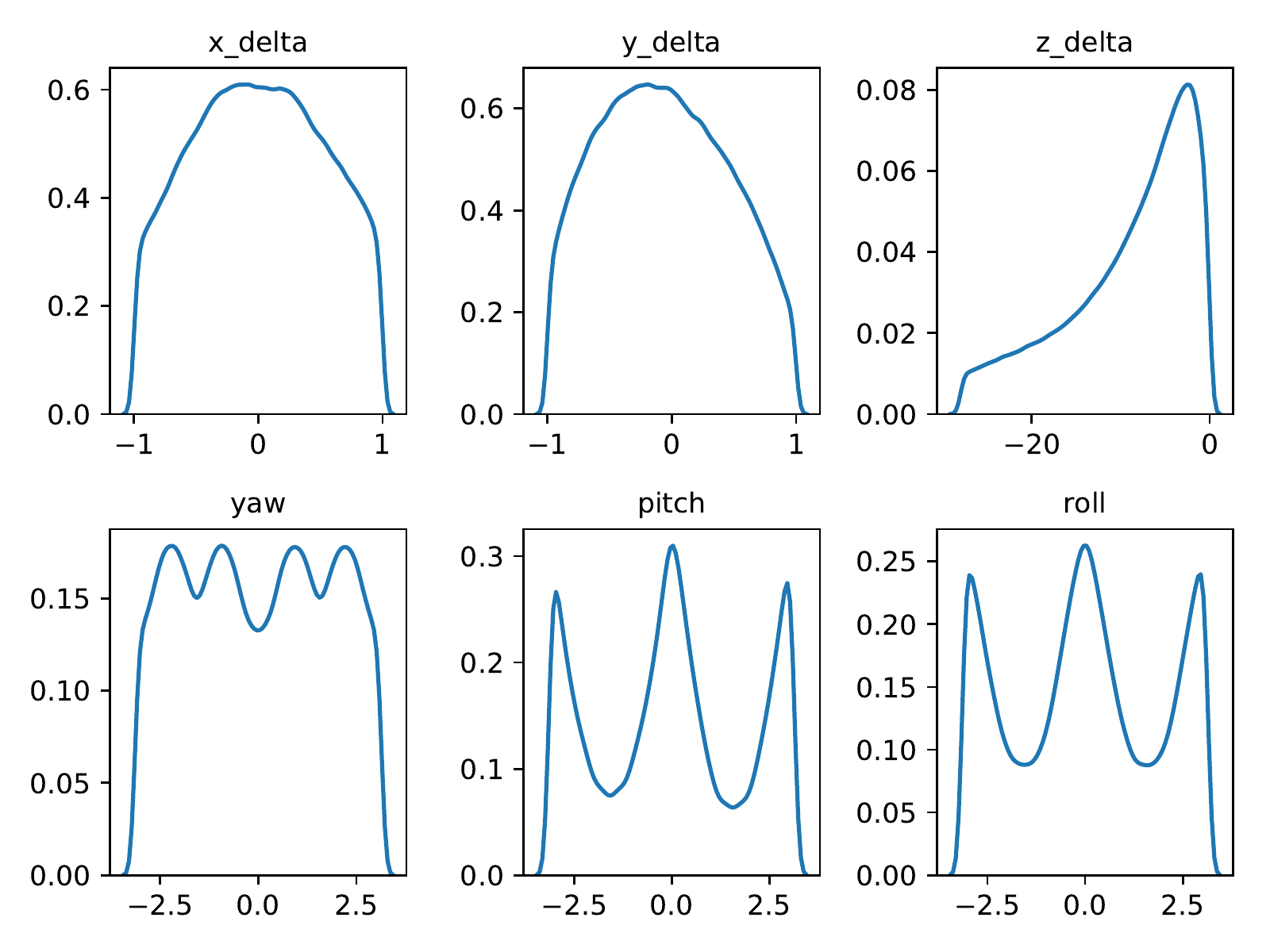}
    \caption{Correct classifications}\label{fig:high_conf_correct_params}
\end{subfigure}
\caption{The distributions of individual pose parameters for (a) high-confidence ($p \geq 0.7$) incorrect classifications and (b) correct classifications obtained from the random sampling procedure described in Sec.~\ref{sec:random_search}.
$x_{\delta}$ and $y_{\delta}$ have been normalized w.r.t. their corresponding $s$ from Eq.~\ref{eq:max_trans}.}
\label{fig:param_distributions}
\end{figure}


\subsec{Experiment}
To test DNN robustness to object rotations and translations, we used RS to generate samples for every 3D object in our dataset.
In addition, to explore the impact of lighting on DNN performance, we considered three different lighting settings: \bright, \medium, and \dark (example renders in Fig.~\ref{fig:light_intensity}).
In all three settings, both the directional light and the ambient light were white in color, \ie, had RGB values of $(1.0, 1.0, 1.0)$, and the directional light was oriented at $(0, -1, 0)$ (\ie, pointing straight down).
The directional light intensities and ambient light intensities were $(1.2, 1.6)$, $(0.4, 1.0)$, and $(0.2, 0.5)$ for the \bright, \medium, and \dark settings, respectively.
All other experiments used the \medium lighting setting.

\subsec{Misclassifications uniformly cover the pose space}
For each object, we calculated the DNN accuracy (\ie, percent of correctly classified samples) across all three lighting settings (Table~\ref{tab:sampling_stats}).
%
The DNN was wrong for the vast majority of samples, \ie, the median percent of correct classifications for all 30 objects was only 3.09\%.
We verified the discovered adversarial poses transfer to the real world by using the 3D objects to reproduce natural, misclassified poses found on the Internet (see Sec.~\ref{sec:transfer}).
High-confidence misclassifications ($p \geq 0.7$) are largely uniformly distributed across every pose parameter (Fig.~\ref{fig:high_conf_params}), \ie, AXs can be found throughout the parameter landscape (see Fig.~\ref{fig:30_ax} for examples).
In contrast, correctly classified examples are highly multimodal w.r.t. the rotation axis angles and heavily biased towards $z_{\delta}$ values that are closer to the camera (Fig.~\ref{fig:high_conf_correct_params}; also compare Fig.~\ref{fig:tsne_img_correct} vs. Fig.~\ref{fig:tsne_img_ax}).
Intriguingly, for ball-like objects (not included in our main traffic dataset), the DNN was far more accurate across the pose space (see Sec.~\ref{sup:ball_objects}).


\subsec{An object can be misclassified as many different labels}
Previous research has shown that it is relatively easy to produce AXs corresponding to many different classes when optimizing input images~\cite{szegedy2013intriguing} or 3D object textures~\cite{Athalye2017}, which are very high-dimensional.
When finding adversarial poses, one might expect---because all renderer parameters, including the original object geometry and textures, are held constant---the success rate to depend largely on the similarities between a given 3D object and examples of the target in ImageNet.
Interestingly, across our 30 objects, RS discovered $990/1000$ different ImageNet classes (132 of which were shared between all objects).
When only considering high-confidence ($p \geq 0.7$) misclassifications, our 30 objects were still misclassified into $797$ different classes with a median number of 240 incorrect labels found per object (see Fig.~\ref{fig:common_failures_per_label} and Fig.~\ref{fig:tsne_img_ax} for examples).
Across all adversarial poses and objects, DNNs tend to be more confident when correct than when wrong (the median of median probabilities were 0.41 vs. 0.21, respectively).

\begin{figure}[h]
\begin{subfigure}{\linewidth}
    \centering
    \includegraphics[width=0.92\linewidth]{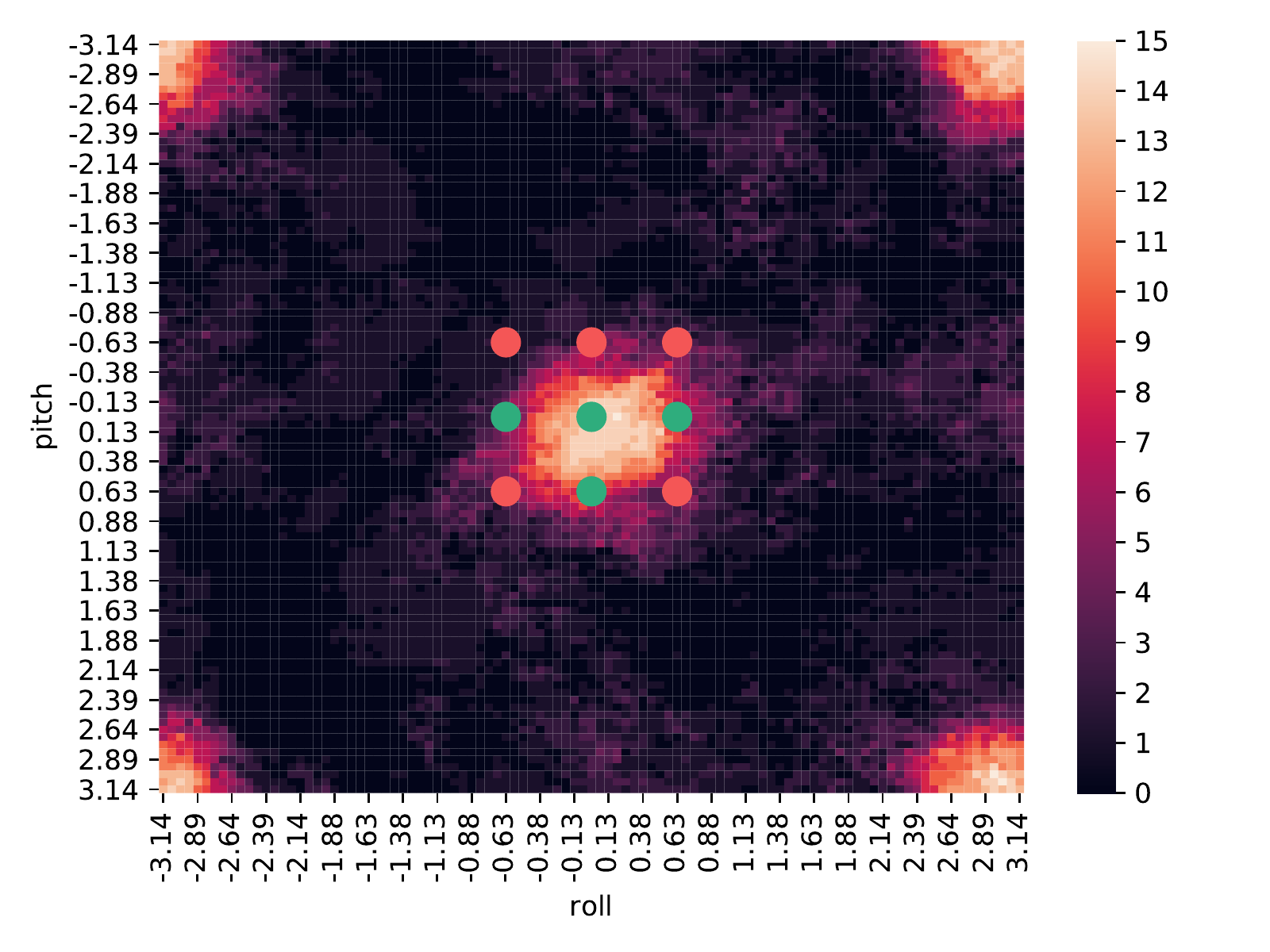}
    \caption{}\label{fig:fire_truck_roll_pitch}
\end{subfigure}
\begin{subfigure}{\linewidth}
    \centering
    \includegraphics[width=0.75\linewidth]{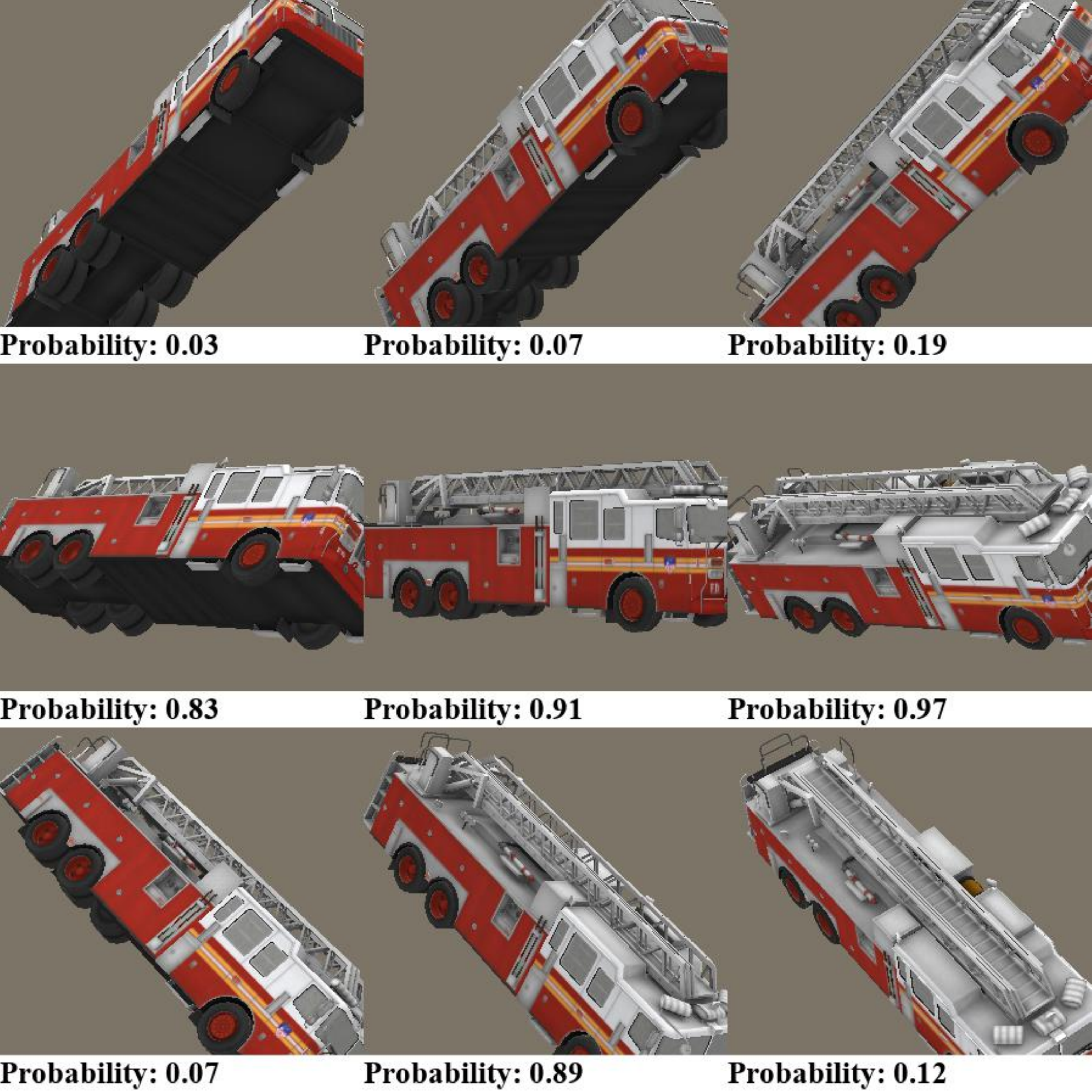}
    \caption{}\label{fig:fire_truck_pitch_roll_collage}
\end{subfigure}
\caption{
Inception-v3's ability to correctly classify images is highly localized in the rotation and translation parameter space.
(a) The classification landscape for 15 vehicle objects when altering $\theta_{r}$ and $\theta_{p}$ and holding $(x_{\delta}, y_{\delta}, z_{\delta}, \theta_{y})$ at $(0, 0, -3, \frac{\pi}{4})$.
Lighter regions correspond to poses with a greater number of correctly classified vehicle objects.
Green and red circles indicate correct and incorrect classifications, respectively, corresponding to the fire truck object poses found in (b).
}\label{fig:landscape}
\end{figure}

\subsection{Common object classifications are shared across different lighting settings}



Here, we analyze how our results generalize across different lighting conditions.
From the data produced in Sec.~\ref{sec:easily_confused}, for each object, we calculated the
DNN accuracy under each lighting setting.
Then, for each object, we took the absolute difference of the accuracies for all three lighting combinations (\ie, \bright vs.~\medium, \bright vs.~\dark, and \medium vs.~\dark) and recorded the maximum of those values.
The median ``maximum absolute difference'' of accuracies for all objects was 2.29\% (compared to the median accuracy of $3.09\%$ across all lighting settings).
That is, DNN accuracy is consistently low across all lighting conditions.
Lighting changes would not alter the fact that DNNs are vulnerable to adversarial poses.

We also recorded the 50 most frequent classes for each object under the different lighting settings ($S_{b}$, $S_{m}$, and $S_{d}$).
Then, for each object, we computed the intersection over union score $o_{S}$ for these sets:

\vspace*{-0.3cm}
\begin{equation}
o_{S} = 100 \cdot \frac{|S_{b} \cap S_{m} \cap S_{d}|}{|  S_{b} \cup S_{m} \cup S_{d} |}
\end{equation}

%
%
%
%

\noindent
The median $o_{S}$ for all objects was 47.10\%.
That is, for 15 out of 30 objects, 47.10\% of the 50 most frequent classes were shared across lighting settings.
While lighting does have an impact on DNN misclassifications (as expected), the large number of shared labels across lighting settings suggests ImageNet classes are strongly associated with certain adversarial poses regardless of lighting.

\subsection{Correct classifications are highly localized in the rotation and translation landscape}
\label{sec:landscape}

To gain some intuition for how Inception-v3 responds to rotations and translations of an object, we plotted the probability and classification landscapes for paired parameters (\eg, Fig.~\ref{fig:landscape}; pitch vs. roll) while holding the other parameters constant.
We qualitatively observed that the DNN's ability to recognize an object (\eg, a fire truck) in an image varies radically as the object is rotated in the world (Fig.~\ref{fig:landscape}).
Further, adversarial poses often generalize across similar objects (\eg, 83\% of the sampled poses were misclassified for \textit{all} 15 four-wheeled vehicle objects).


\subsec{Experiment}
To quantitatively evaluate the DNN's sensitivity to rotations and translations, we tested how it responded to single parameter disturbances.
For each object, we randomly selected 100 distinct starting poses that the DNN had correctly classified in our random sampling runs.
Then, for each parameter (\eg, yaw rotation angle), we randomly sampled 100 new values\footnote{\label{note:sampling}using the random sampling procedure described in Sec.~\ref{sec:random_search}} while holding the others constant.
For each sample, we recorded whether or not the object remained correctly classified, and then computed the failure (\ie, misclassification) rate for a given (object,~parameter) pair.
Plots of the failure rates for all (object,~parameter) combinations can be found in Fig.~\ref{fig:sensitivity}.

Additionally, for each parameter, we calculated the median of the median failure rates.
That is, for each parameter, we first calculated the median failure rate for all objects, and then calculated the median of those medians for each parameter.
Further, for each (object,~starting pose,~parameter) triple, we recorded the magnitude of the smallest parameter change that resulted in a misclassification.
Then, for each (object,~parameter) pair, we recorded the median of these minimum values.
Finally, we again calculated the median of these medians across objects (Table~\ref{tab:fail_rates}).

\subsec{Results}
As can be seen in Table~\ref{tab:fail_rates}, the DNN is highly sensitive to all single parameter disturbances, but it is especially sensitive to disturbances along the depth ($z_{\delta}$), pitch ($\theta_{p}$), and roll ($\theta_{r}$).
To aid in the interpretation of these results, we converted the raw disturbance values in Table~\ref{tab:fail_rates} to image units.
For $x_{\delta}$ and $y_{\delta}$, the interpretable units are the number of pixels the object shifted in the $x$ or $y$ directions \textit{of the image} (however, note that 3D translations are \textit{not} equivalent to 2D translations due to the perspective projection).

We found that a change in rotation as small as $8.02\degree$ can cause an object to be misclassified (Table~\ref{tab:fail_rates}).
Along the spatial dimensions, a translation resulting in the object moving as few as $2$ px horizontally or $4.5$ px vertically also caused the DNN to misclassify.\footnote{Note that the sensitivity of classifiers and object detectors to \textit{2D} translations has been observed in concurrent work \cite{rosenfeld2018elephant,engstrom2019a,zhang2019making,azulay2018deep}.}
Lastly, along the $z$-axis, a change in ``size'' (\ie, the area of the object's bounding box) of only 5.4\% can cause an object to be misclassified.

\begin{table}[h]
	\begin{center}
		\begin{tabular}{cccc}
			\toprule
			Parameter & Fail Rate (\%) & Min. $\Delta$ & Int. $\Delta$ \\
			\midrule
			$x_{\delta}$ & 42 & 0.09 & 2.0 px \\
			$y_{\delta}$ & 49 & 0.10 & 4.5 px \\
			$z_{\delta}$ & 81 & 0.77 & 5.4\% \\
			$\theta_{y}$ & 69 & 0.18 & $10.31\degree$ \\
			$\theta_{p}$ & 83 & 0.14 & $8.02\degree$ \\
			$\theta_{r}$ & 81 & 0.16 & $9.17\degree$ \\
			\bottomrule
		\end{tabular}
	\end{center}
	\caption{The median of the median failure rates and the median of the median minimum disturbances (Min. $\Delta$) for the single parameter sensitivity tests described in Section \ref{sec:landscape}.
	Int. $\Delta$ converts the values in Min. $\Delta$ to more interpretable units.
	For $x_{\delta}$ and $y_{\delta}$, the interpretable units are pixels.
	For $z_{\delta}$, the interpretable unit is the percent change in the area of the bounding box containing the object.
		See main text and Fig.~\ref{fig:sensitivity} for additional information.
	}
	\label{tab:fail_rates}
\end{table}

\subsection{Optimization methods can effectively generate targeted adversarial poses}
\label{sec:comparing_methods}

Given a challenging, highly non-convex objective landscape (Fig.~\ref{fig:landscape}), we wish to evaluate the effectiveness of two different types of approximate gradients at targeted attacks, \ie, finding adversarial examples misclassified as a target class \cite{szegedy2013intriguing}.
Here, we compare (1) random search; (2) gradient descent with finite-difference gradients (FD-G); and (3) gradient descent with analytical, approximate gradients provided by a differentiable renderer (DR-G) \cite{kato2018neural}.

\subsec{Experiment}
Because our adversarial pose attacks are inherently constrained by the fixed geometry and appearances of a given 3D object (see Sec.~\ref{sec:easily_confused}),
we defined the targets to be the 50 most frequent incorrect classes found by our RS procedure for each object.
For each (object, target) pair, we ran 50 optimization trials using ZRS, FD-G, and DR-G.
All treatments were initialized with a pose found by the ZRS procedure and then allowed to optimize for 100 iterations.

\subsec{Results}
For each of the 50 optimization trials, 
we recorded both whether or not the target was hit and the maximum target probability obtained during the run.
For each (object,~target) pair, we calculated the percent of target hits and the median maximum confidence score of the target labels (see Table~\ref{tab:optim_stats}).
As shown in Table~\ref{tab:optim_stats}, FD-G is substantially more effective than ZRS at generating targeted adversarial poses, having both higher median hit rates and confidence scores.
In addition, we found the approximate gradients from DR to be surprisingly noisy, and DR-G largely underperformed even non-gradient methods (ZRS) (see Sec.~\ref{sec:kr}).

\begin{table}[h]
  \centering
  \begin{tabular}{lcrr}
    \toprule
    & Hit Rate (\%) & Target Prob. \\
    \midrule
    ZRS ~~~~random search & 78 & 0.29 \\
    FD-G ~~gradient-based & \textbf{92} & \textbf{0.41} \\
    DR-G\(^\dagger\) gradient-based & 32 & 0.22 \\
    \bottomrule
  \end{tabular}
  \caption{The median percent of target hits and the median of the median target probabilities
  	for random search (ZRS), gradient descent with finite difference gradients (FD-G), and DR gradients (DR-G).
  	All attacks are targeted and initialized with $z_{\delta}$-constrained random search.
  \(^\dagger\)DR-G is not directly comparable to FD-G and ZRS
  (details in Sec.~\ref{sec:dr-expr}).
	}
  \label{tab:optim_stats}
\end{table}

\subsection{Adversarial poses transfer to different image classifiers and object detectors}

The most important property of previously documented AXs is that they transfer across ML models, enabling black-box attacks~\cite{Yuan2017}.
Here, we investigate the transferability of our adversarial poses to (a) two different image classifiers, AlexNet~\cite{Krizhevsky2012} and ResNet-50~\cite{He2016}, trained on the same ImageNet dataset; and (b) an object detector YOLOv3~\cite{Redmon2018} trained on the MS COCO dataset~\cite{lin2014microsoft}.



For each object, we randomly selected 1,350 AXs that were misclassified by Inception-v3 with high confidence ($p \geq 0.9$) from our untargeted RS experiments in Sec.~\ref{sec:easily_confused}.
We exposed the AXs to AlexNet and ResNet-50 and calculated their misclassification rates.
We found that almost all AXs transfer with median misclassification rates of 99.9\% and 99.4\% for AlexNet and ResNet-50, respectively.
In addition, 10.1\% of AlexNet misclassifications and 27.7\% of ResNet-50 misclassifications were identical to the Inception-v3 predicted labels.

There are two orthogonal hypotheses for this result. 
First, the ImageNet training-set images themselves may contain a strong bias towards common poses, omitting uncommon poses (Sec.~\ref{sec:nearest_neighbors} shows supporting evidence from a nearest-neighbor test). 
Second, the models themselves may not be robust to even slight disturbances of the known, in-distribution poses.



\subsec{Object detectors}
Previous research has shown that object detectors can be more robust to adversarial attacks than image classifiers~\cite{lu2017-standard}.
Here, we investigate how well our AXs transfer to a state-of-the-art object detector---YOLOv3.
YOLOv3 was trained on MS COCO, a dataset of bounding boxes corresponding to 80 different object classes.
We only considered the 13 objects that belong to classes present in both the ImageNet and MS COCO datasets.
We found that 75.5\% of adversarial poses generated for Inception-v3 are also misclassified by YOLOv3 (see Sec.~\ref{sec:yolo} for more details).
These results suggest the adversarial pose problem transfers across datasets, models, and tasks.


\subsection{Adversarial training}
One of the most effective methods for defending against OoD examples has been adversarial training \cite{Goodfellow2014}, \ie augmenting the training set with AXs---also a common approach in anomaly detection \cite{chandola2009anomaly}.
We tested whether adversarial training can improve DNN robustness to new poses generated for (1) our 30 training-set 3D objects; and (2) seven held-out 3D objects (see Sec.~\ref{sec:adversarial_training} for details).
Following adversarial training, the accuracy of the DNN substantially increased for \emph{known} objects (Table~\ref{tab:ax_stats}; $99.67\%$ vs. $6.7\%$).
However, the model (AT) still misclassified the adversarial poses of held-out objects at an 89.2\% error rate.

\begin{table}[h]
\begin{center}
  \begin{tabular}{lcc}
    \toprule
     & PT & AT \\
    \midrule
    Error (T) & 99.67 & 6.7 \\
    Error (H) & 99.81 & 89.2 \\
    \midrule
    High-confidence Error (T) & 87.8 & 1.9 \\        High-confidence Error (H) & 48.2 & 33.3 \\
    \bottomrule
\end{tabular}
\end{center}
\caption{The median percent of misclassifications (Error) and high-confidence (\ie, $p > 0.7$) misclassifications by the pre-trained AlexNet (PT) and our AlexNet trained with adversarial examples (AT) on random poses of training-set objects (T) and held-out objects (H).}
\label{tab:ax_stats}
\end{table}

\section{Related work}



\subsec{Out-of-distribution detection}
OoD classes, \ie, classes not found in the training set, present a significant challenge for computer vision technologies in real-world settings~\cite{scheirer2013toward}.
Here, we study an orthogonal problem---correctly classifying OoD poses of objects from \emph{known} classes.
While rejecting to classify is a common approach for handling OoD examples~\cite{hendrycks2016baseline,scheirer2013toward}, the OoD poses 
in our work
come from known classes and thus \emph{should be} assigned correct labels.


\subsec{2D adversarial examples} Numerous techniques for crafting AXs that fool image classifiers have been discovered~\cite{Yuan2017}.
However, previous work has typically optimized in the 2D input space~\cite{Yuan2017}, \eg, by synthesizing an entire image~\cite{nguyen2015deep}, a small patch~\cite{karmon2018lavan,evtimov2017robust}, a few pixels~\cite{carlini2017towards}, or only a single pixel~\cite{su2017one}.
But pixel-wise changes are uncorrelated~\cite{nguyen2017plug}, so pixel-based attacks may not transfer well to the real world~\cite{Lu2017,Luo2015} because there is an infinitesimal chance that such specifically crafted, uncorrelated pixels will be encountered in the vast physical space of camera, lighting, traffic, and weather configurations.
\cite{xiao2018spatially} generated spatially transformed adversarial examples that are perceptually realistic and more difficult to defend against, but the technique still directly operates on pixels.

\subsec{3D adversarial examples}
Athalye et al.~\cite{Athalye2017} used a 3D renderer to synthesize textures for a 3D object such that, under a wide range of camera views, the object was still rendered into an effective AX.
We also used 3D renderers, but instead of optimizing textures, we optimized the poses of known objects to cause DNNs to misclassify (\ie, we kept the textures, lighting, camera settings, and background image constant).

\subsec{Concurrent work}
We describe below two concurrent attempts that are closely related to ours.
First, Liu et al.~\cite{Liu2018} proposed a differentiable 3D renderer and used it to perturb both an object's geometry and the scene's lighting to cause a DNN to misbehave.
However, their geometry perturbations were constrained to be infinitesimal so that the visibility of the vertices would not change.
Therefore, their result of minutely perturbing the geometry is effectively similar to that of perturbing textures~\cite{Athalye2017}.
In contrast, we performed 3D rotations and 3D translations to move an object inside a 3D space (\ie, the viewing frustum of the camera).

Second, Engstrom et al.~\cite{engstrom2019a} showed how simple 2D image rotations and translations can cause DNNs to misclassify.
However, these 2D transformations still do not reveal the type of adversarial poses discovered by rotating 3D objects (\eg, a flipped-over school bus; Fig.~\ref{fig:teaser}d).

To the best of our knowledge, our work is the first attempt to harness 3D objects to study the OoD poses of well-known training-set objects that cause state-of-the-art ImageNet classifiers and MS COCO detectors to misclassify.

\section{Discussion and conclusion}

In this paper, we revealed how DNNs' understanding of objects like ``school bus'' and ``fire truck'' is quite naive---they can correctly label only a small subset of the entire pose space for 3D objects.
Note that we can also find real-world OoD poses by simply taking photos of real objects (Sec.~\ref{sec:transfer}).
%
We believe classifying an arbitrary pose into one of the object classes is an ill-posed task, and that the adversarial pose problem might be alleviated via multiple orthogonal approaches.
The first is addressing biased data \cite{torralba2011unbiased}.
Because ImageNet and MS COCO datasets are constructed from photographs taken by people, the datasets reflect the aesthetic tendencies of their captors.
Such biases can be somewhat alleviated through data augmentation, specifically, by harnessing images generated from 3D renderers \cite{shrivastava2017learning,alhaija2018geometric}.
From the modeling view, we believe DNNs would benefit from the incorporation of 3D information, \eg, \cite{alhaija2018geometric}.

Finally, our work introduced a new promising method (Fig.~\ref{fig:concept}) for testing computer vision DNNs by harnessing 3D renderers and 3D models.
While we only optimize a single object here, the framework could be extended to jointly optimize lighting, background image, and multiple objects, all in one ``adversarial world''.
Not only does our framework enable us to enumerate test cases for DNNs, but it also serves as an interpretability tool for extracting useful insights about these black-box models' inner functions.

\subsection*{Acknowledgements}
We thank Hiroharu Kato and Nikos Kolotouros for their valuable discussions and help with the differentiable renderer.
We also thank Rodrigo Sardinas for his help with some GPU servers used in the project.
AN is supported by multiple funds from Auburn University, a donation from Adobe Inc., and computing credits from Amazon AWS.

{\small
\bibliographystyle{style/ieee}
\bibliography{references}
}

\clearpage

\renewcommand{\thesection}{S\arabic{section}}
\renewcommand{\thesubsection}{\thesection.\arabic{subsection}}

\newcommand{\beginsupplementary}{%
            \setcounter{table}{0}
    \renewcommand{\thetable}{S\arabic{table}}%
            \setcounter{figure}{0}
    \renewcommand{\thefigure}{S\arabic{figure}}%
    \setcounter{section}{0}
}
\newcommand{\suptitle}{Supplementary materials for:\\\papertitle}

\newcommand{\toptitlebar}{
    \hrule height 4pt
    \vskip 0.25in
    \vskip -\parskip%
}
\newcommand{\bottomtitlebar}{
    \vskip 0.29in
    \vskip -\parskip%
    \hrule height 1pt
    \vskip 0.09in%
}

\beginsupplementary%

\newcommand{\maketitlesupp}{
    \newpage
    \onecolumn
        \null%
        \vskip .375in
        \begin{center}
            {\Large \bf \suptitle\par}
            \vspace*{24pt}
            {
                \large
                \lineskip=.5em
                \par
            }
            \vskip .5em
            \vspace*{12pt}
        \end{center}
}

\maketitlesupp%


\section{Extended description of the 3D object dataset and its evaluation}
\label{sec:SI_3d_object_dataset}

\subsection{Dataset construction}\label{sec:construction}

\begin{figure*}
    \centering
    \includegraphics[width=0.9\columnwidth]{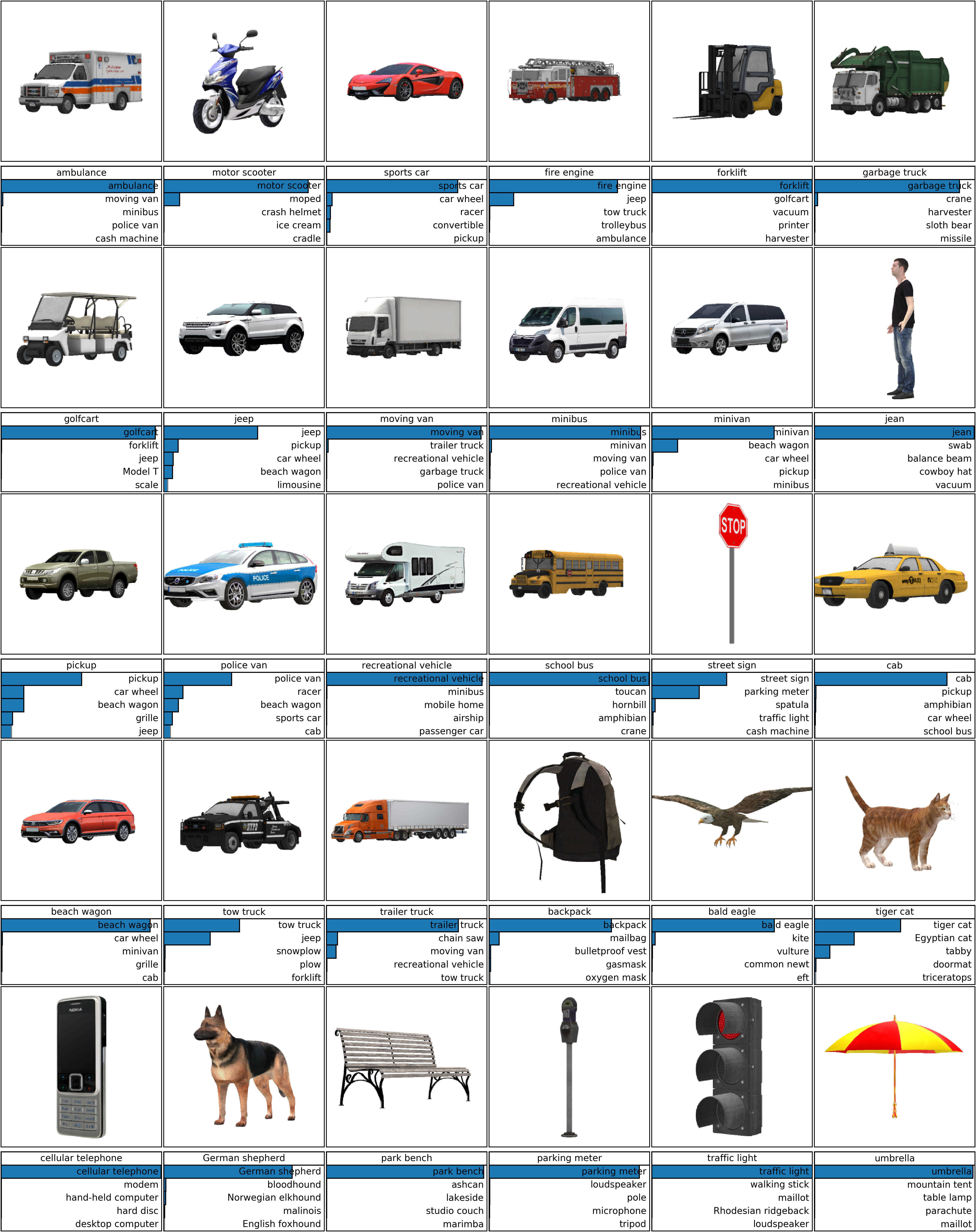}
    \caption{    	
    	We tested Inception-v3's predictions on the renders generated by the differentiable renderer (DR).
    	We show here the top-5 predictions for one random pose per object. 
    	However, in total, we generated 36 poses for each object by (1) varying the object distance to the camera; and (2) rotating the object around the yaw axis.
    	See \url{https://goo.gl/7LG3Cy} for all the renders and DNN top-5 predictions.
    	Across all 30 objects, on average, Inception-v3 correctly recognizes 83.2\% of the renders.
    	See Sec.~\ref{sec:quantitative_eval} for more details.
    }\label{fig:dataset_A}
\end{figure*}

\subsec{Classes}
Our main dataset consists of 30 unique 3D object models corresponding to 30 ImageNet classes relevant to a traffic environment.
The 30 classes include 20 vehicles (\eg, \class{school bus} and \class{cab}) and 10 street-related items (\eg, \class{traffic light}). 
See Fig.~\ref{fig:dataset_A} for example renders of each object.

\subsec{Acquisition}
We collected 3D objects and constructed our own datasets for the study.
3D models with high-quality image textures were purchased from \url{turbosquid.com}, \url{free3d.com}, and \url{cgtrader.com}.

To make sure the renders were as close to real ImageNet photos as possible, we used only 3D models that had high-quality 2D image textures.
We did not choose 3D models from public datasets, \eg, ObjectNet3D \cite{xiang2016objectnet3d}, because most of them do not have high-quality image textures.
While the renders of such models may be correctly classified by DNNs, we excluded them from our study because of their poor realism.
We also examined the ImageNet images to ensure they contained real-world examples qualitatively similar to each 3D object in our 3D dataset.

\subsec{3D objects}
Each 3D object is represented as a mesh, \ie, a list of triangular faces, each defined by three vertices \cite{marschner2015fundamentals}.
The 30 meshes have on average $9,908$ triangles (see Table~\ref{tab:num_triangles} for specific numbers).


  \begin{table}[h]
	\centering
	\begin{tabular}{lrr}
		\toprule
		3D object        & Tessellated \(N_T\) & Original \(N_O\) \\
		\midrule
		\class{ambulance}            & 70,228 & 5,348 \\
		\class{backpack}             & 48,251 & 1,689\\
		\class{bald eagle}           & 63,212 & 2,950\\
		\class{beach wagon}          & 220,956 & 2,024\\
		\class{cab}                  & 53,776 & 4,743\\
		\class{cellphone}   & 59,910 & 502\\
		\class{fire engine}          & 93,105 & 8,996\\
		\class{forklift}             & 130,455 & 5,223\\
		\class{garbage truck}        & 97,482 & 5,778\\
		\class{German shepherd}      & 88,496 & 88,496\\
		\class{golf cart}            & 98,007 & 5,153\\
		\class{jean}                 & 17,920 & 17,920\\
		\class{jeep}                 & 191,144 & 2,282\\
		\class{minibus}              & 193,772 & 1,910\\
		\class{minivan}              & 271,178 & 1,548\\
		\bottomrule
	\end{tabular}
	\begin{tabular}{lrr}
		\toprule
		3D object        & Tessellated \(N_T\) & Original \(N_O\) \\
		\midrule
		\class{motor scooter}        & 96,638 & 2,356\\
		\class{moving van}           & 83,712 & 5,055\\
		\class{park bench}           & 134,162 & 1,972\\
		\class{parking meter}        & 37,246 & 1,086\\
		\class{pickup}               & 191,580 & 2,058\\
		\class{police van}           & 243,132 & 1,984\\
		\class{recreational vehicle} & 191,532 & 1,870\\
		\class{school bus}           & 229,584 & 6,244\\
		\class{sports car}           & 194,406 & 2,406\\
		\class{street sign}          & 17,458 & 17,458\\
		\class{tiger cat}            & 107,431 & 3,954\\
		\class{tow truck}            & 221,272 & 5,764\\
		\class{traffic light}        & 392,001 & 13,840\\
		\class{trailer truck}        & 526,002 & 5,224\\
		\class{umbrella}             & 71,410 & 71,410\\
		\bottomrule
	\end{tabular}
	\caption{The triangle number for the 30 objects used in our study.  
		\(N_O\) shows the number of
		triangles for the original 3D objects, and 		\(N_T\) shows the same number after tessellation.
		  Across 30 objects, the average triangle count increases $\sim15$x from $\overline{N_O} = 9,908$ to		  
		  \(\overline{N_T} = 147,849\).}\label{tab:num_triangles}
\end{table}

\subsection{Manual object tessellation for experiments using the Differentiable Renderer}

In contrast to ModernGL \cite{modernGL}---the non-differentiable renderer (NR) in our paper---the differentiable renderer (DR) by Kato et. al~\cite{kato2018neural} does not perform tessellation, a standard process to increase the resolution of renders.
Therefore, the render quality of the DR is lower than that of the NR. 
To minimize this gap and make results from the NR more comparable with those from the DR, we manually tessellated each 3D object as a pre-processing step for rendering with the DR.
Using the manually tessellated objects, we then (1) evaluated the render quality of the DR (Sec.~\ref{sec:evaluation}); and (2) performed research experiments with the DR (\ie, the DR-G method in Sec.~\ref{sec:comparing_methods}).\\

\begin{figure}[h]
	\begin{center}
		\includegraphics[width=0.7\linewidth]{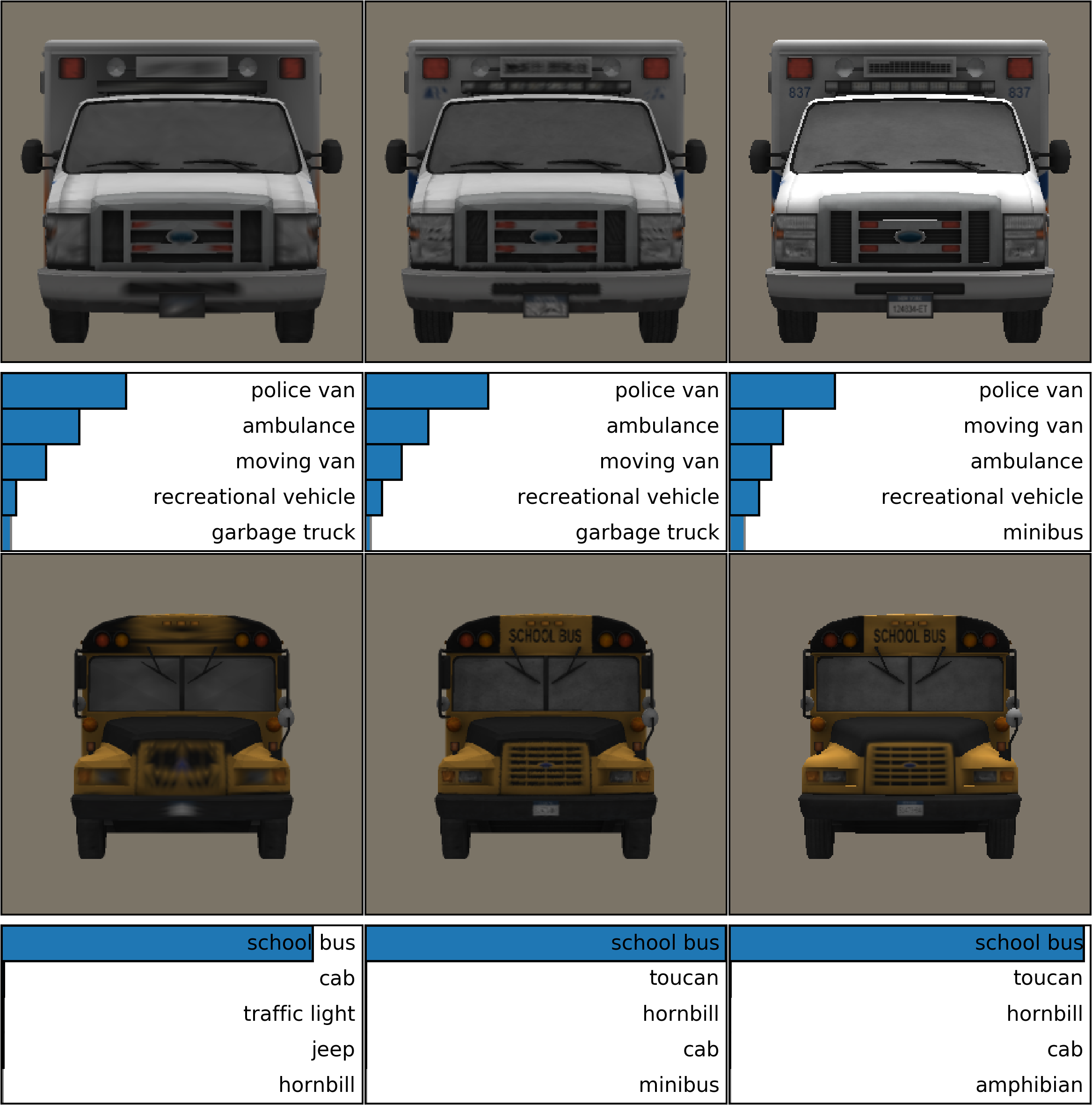}
	\end{center}
	\vspace*{-0.3cm}
	\centering
{
	\hspace{2.7cm}
	(a) DR without tessellation
	\hspace{0.35cm} (b) DR with tessellation
	\hspace{0.4cm} (c) NR with tessellation
	\hfill
}
	\caption{A comparison of 3D object renders (here, \class{ambulance} and \class{school bus}) before and after tessellation.
		\\(a) Original 3D models rendered by the differentiable renderer (DR) \cite{kato2018neural} without tessellation.
		\\(b) DR renderings of the same objects after manual tessellation.
		\\(c) The non-differentiable renderer (NR), \ie, ModernGL \cite{modernGL}, renderings of the original objects.
		\\After manual tessellation, the render quality of the DR appears to be sharper (a vs. b) and closely matches that of the NR, which also internally tessellates objects (b vs. c).	
		}
	
	\label{fig:compare_tessellation}
\end{figure}

\subsec{Tessellation} 
We used the \emph{Quadify Mesh Modifier} feature (quad size of 2\%) in 3ds~Max 2018 to tessellate objects, increasing the average number of faces $\sim$15x from $9,908$ to $147,849$ (see Table~\ref{tab:num_triangles}).
The render quality after tessellation is sharper and of a higher resolution (see Fig.~\ref{fig:compare_tessellation}a vs. b).
Note that the NR pipeline already performs tessellation for every input 3D object.
Therefore, we did not perform manual tessellation for 3D objects rendered by the NR.



%


\subsection{Evaluation}
\label{sec:evaluation}

We recognize that a reality gap will often exist between a render and a real photo.
Therefore, we rigorously evaluated our renders to make sure the reality gap was acceptable for our study.
From $\sim$100 initially-purchased 3D object models, we selected the 30 highest-quality objects that both (1) passed a visual human Turing test; and (2) were correctly recognized with high confidence by the Inception-v3 classifier~\cite{szegedy2016rethinking}.

\subsubsection{Qualitative evaluation} 
\label{sec:qualitative_eval}

Here, we attempt to provide a qualitative ``apples to apples'' comparison between renders of our high-quality 3D objects and photos of their real-world counterparts by generating (real photo, render) pairs. 
The entire process follows the standard pose-annotation procedure (\eg, for the Pix3D \cite{pix3d} or YCB-Video \cite{xiang2017posecnn} datasets) and is described below:

\begin{enumerate}
	\item We retrieved $\sim$3 real photos for each 3D object (\eg, a car) from the Internet using descriptive information (\eg, a car's make, model, and year).
	\item For each real photo, we replaced the object with matching background content via Adobe Photoshop's Context-Aware Fill-In feature to obtain a background-only (\ie, no foreground objects) photo $B$.
	\item We then rendered the 3D object on the background $B$ obtained in Step 2 and manually aligned the pose of the 3D object so that it closely matched the reference photo.
	\item Finally, we evaluated the (photo, render) pairs in a side-by-side comparison.
\end{enumerate}


\noindent In total, we generated  116 (photo, render) pairs for two sets of 3D objects:
\begin{itemize}
	\item Set 1: The 30 objects used as the 3D dataset in our main experiments (see Fig.~\ref{fig:dataset_B}).
	All $30$ objects $\times$ $2$ pairs $= 60$ pairs are provided at \url{https://drive.google.com/drive/folders/1ti9zo1dzU1e9b-mpqv0bhTrMeoeUBULm}.
	The pose alignment was done in our GUI tool\footnote{\url{https://github.com/airalcorn2/strike-with-a-pose}}.
	The scenes were rendered via the NR (\ie, ModernGL).
	\item Set 2: 17 objects gathered separately from Set 1 only for evaluation.
	We collected the 17 extra objects because we were able to find Internet photos of their exact real-world counterparts (\eg, photos of the 2014 Mercedes-Benz E-Klasse Coupe).
	These 17 objects are of the same high quality as the 30 main objects.
	The pose alignment was done in Blender, and the scenes were rendered with the DR.
	All $56$ pairs generated from these $17$ objects are provided at \url{https://goo.gl/8z42zL}.
\end{itemize}


While discrepancies can be visually spotted in our side-by-side comparisons, we found most of the renders passed our human visual Turing test if presented alone.
That is, it is not easy for humans to tell whether a render is a real photo or not (if they are not primed with the reference photos).

\subsubsection{Quantitative evaluation} 
\label{sec:quantitative_eval}



In addition to the qualitative evaluation, we also quantitatively evaluated the Google Inception-v3 \cite{szegedy2016rethinking}'s top-1 accuracy on renders that use either (a) an empty background or (b) real background images.

\subsubsection*{a. Evaluation of the renders of 30 objects on an empty background} 

Because the experiments in the main text used our self-assembled 30-object dataset (Sec.~\ref{sec:construction}), we describe the process and the results of our quantitative evaluation for only those objects.

We rendered the objects on a white background with RGB values of (1.0, 1.0, 1.0), an ambient light intensity of 0.9, and a directional light intensity of 0.5.
For each object, we sampled 36 unique views (common in ImageNet) evenly divided into three sets.
For each set, we set the object at the origin, the up direction to $(0,1,0)$, and the camera position to $(0,0,-z)$ where $z = \{4, 6, 8\}$.
We sampled 12 views per set by starting the object at a $10^\circ{}$ yaw and generating a render at every $30^\circ{}$ yaw-rotation.
Across all objects and all renders, the Inception-v3 top-1 accuracy is $83.23\%$ (comparable to $77.45\%$ on ImageNet images \cite{szegedy2016rethinking}) with a mean top-1 confidence score of $0.78$.
The top-1 and top-5 average accuracy and confidence scores are shown in
Table~\ref{tab:avg_accuracy_30obj}.

\begin{table}[h]
	\centering
	\begin{tabular}{*{5}{lccc}}
		\toprule
		Distance &  4 & 6 & 8 & Average \\ \midrule
		top-1 mean accuracy & 84.2\% & 84.4\%& 81.1\%& 83.2\%\\
		top-5 mean accuracy & 95.3\% & 98.6\%& 96.7\%& 96.9\% \\
		top-1 mean confidence score & 0.77 & 0.80 & 0.76& 0.78 \\ \bottomrule
	\end{tabular}
	\caption{The top-1 and top-5 average accuracy and confidence scores for Inception-v3~\cite{szegedy2016rethinking} on the renders of the 30 objects in our dataset.}\label{tab:avg_accuracy_30obj}
\end{table}

\subsubsection*{b. Evaluation of the renders of test objects on real backgrounds}

\begin{figure*}[t]
	\begin{center}
		\includegraphics[width=\linewidth]{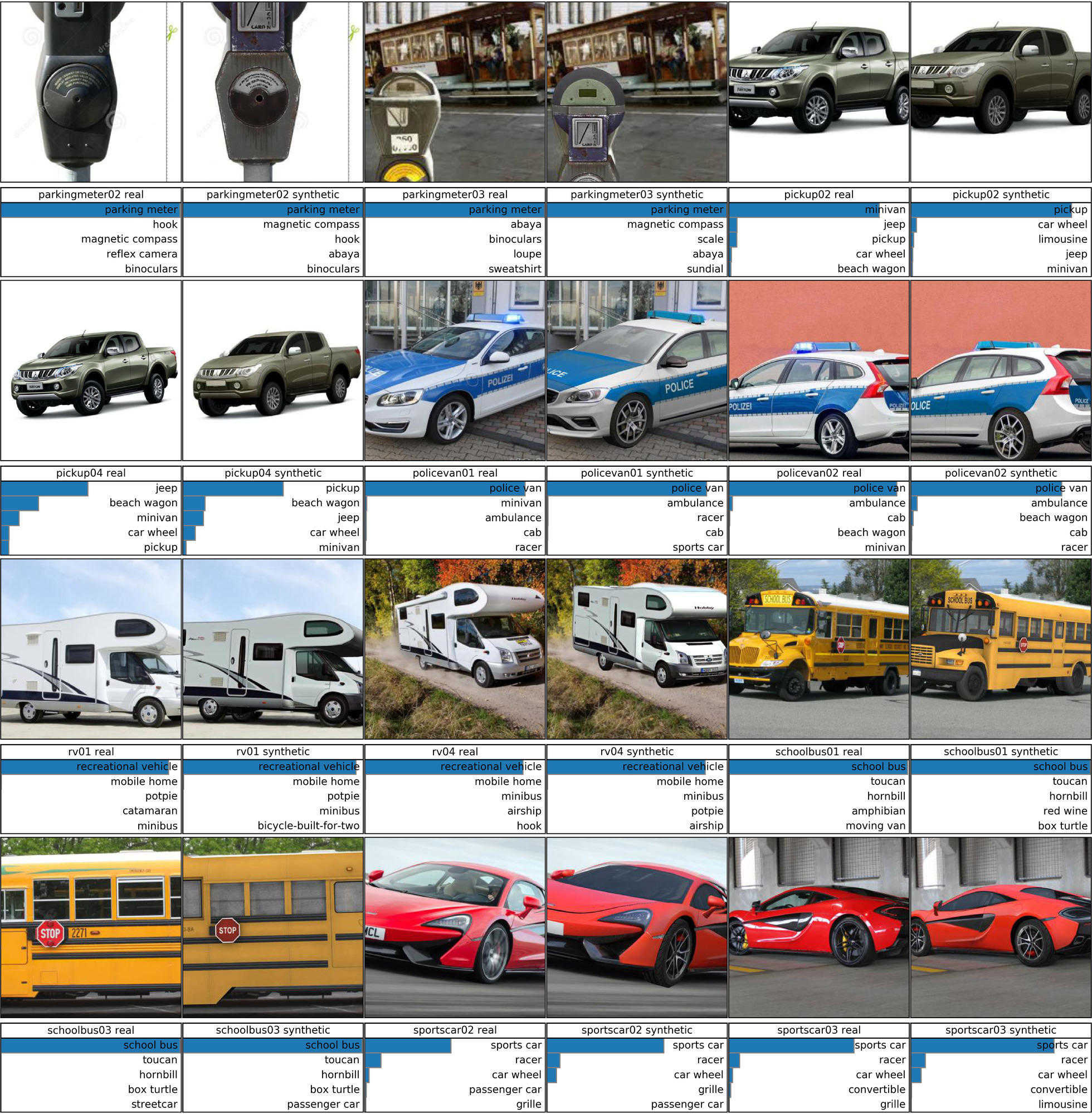}
	\end{center}
	\caption{
		12 random pairs of real photos (left) and renders (right) among 116 pairs produced in total for our 3D object rendering evaluation (Sec.~\ref{sec:qualitative_eval}).
		The renders are produced by ModernGL.
		More comparison images are available at \url{https://drive.google.com/drive/folders/1ti9zo1dzU1e9b-mpqv0bhTrMeoeUBULm}.
		While discrepancies can be spotted in our side-by-side comparisons, we found that most of the renders passed our human visual Turing test if presented alone.
	}
	\label{fig:dataset_B}
\end{figure*}

In addition to our qualitative side-by-side (real photo, render) comparisons (Fig.~\ref{fig:dataset_B}), we quantitatively compared Inception-v3's predictions for our renders to those for real photos.
We found a high similarity between real photos and renders for DNN predictions.
That is, across all 56 pairs (Sec.~\ref{sec:qualitative_eval}), the top-1 predictions match 71.43\% of the time.
Across all pairs, 76.06\% of the top-5 labels for real photos match those for renders.

%
%
%

\begin{figure*}
	\centering
	\includegraphics[width=1.0\columnwidth]{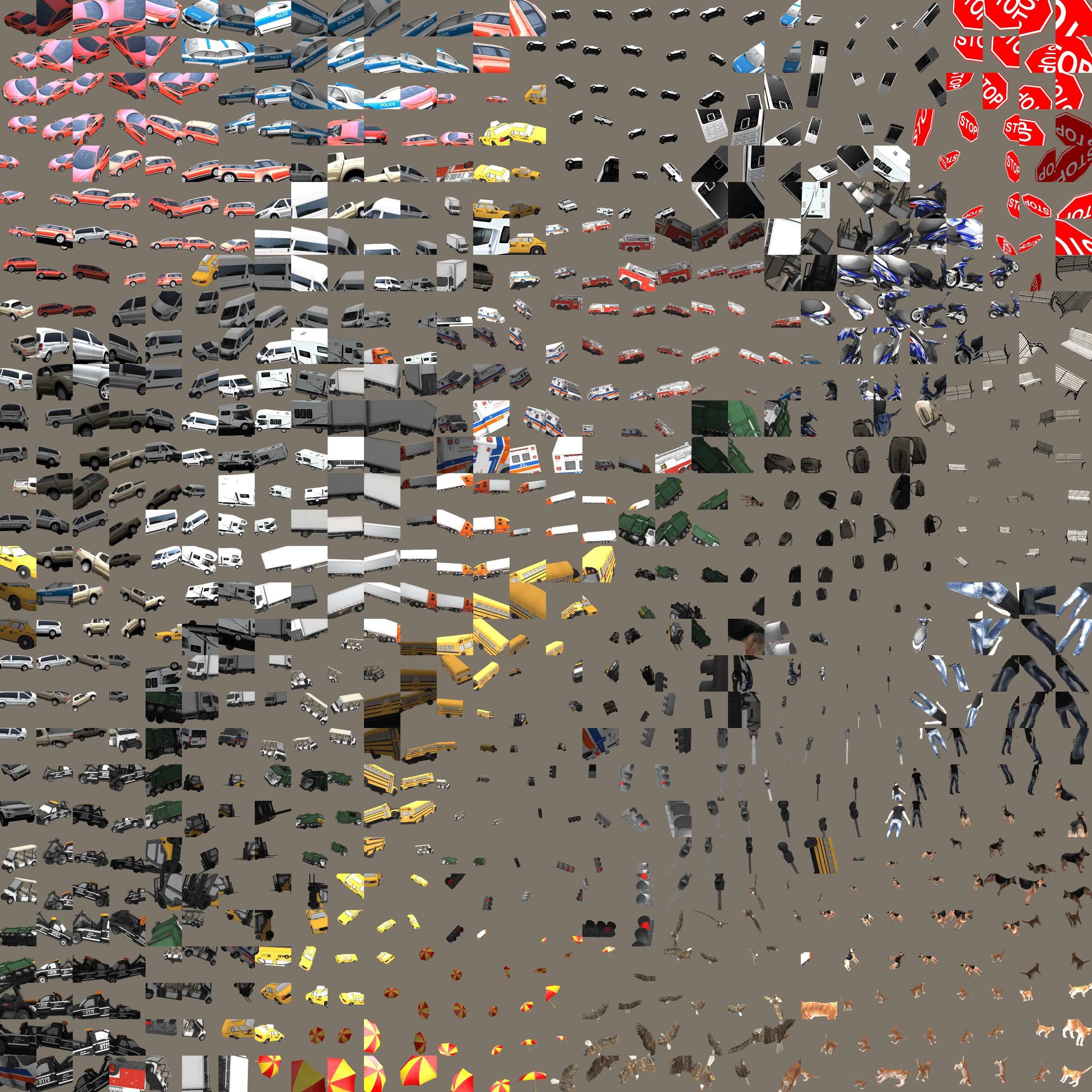}
	\caption{ 
		For each object, we collected 30 high-confidence ($p \geq 0.9$) correctly classified images by Inception-v3.
		The images were generated via the random search procedure.
		We show here a grid t-SNE of AlexNet \cite{Krizhevsky2012} \layer{fc7} features for all $30$ objects $\times$ $30$ images = $900$ images.
		Correctly classified images for each object tend to be similar and clustered together.
		The original, high-resolution figure is available at \url{https://goo.gl/TGgPgB}.
		\\To better visualize the clusters, we plotted the same t-SNE but used unique colors to denote the different 3D objects in the renders  (Fig.~\ref{fig:tsne_color_correct}).
		Compare and contrast this plot with the t-SNE of only misclassified poses 
		(Figs.~\ref{fig:tsne_img_ax} \&~\ref{fig:tsne_color_ax}).
	}\label{fig:tsne_img_correct}
\end{figure*}

\begin{figure*}
	\centering
	\includegraphics[width=1.0\columnwidth]{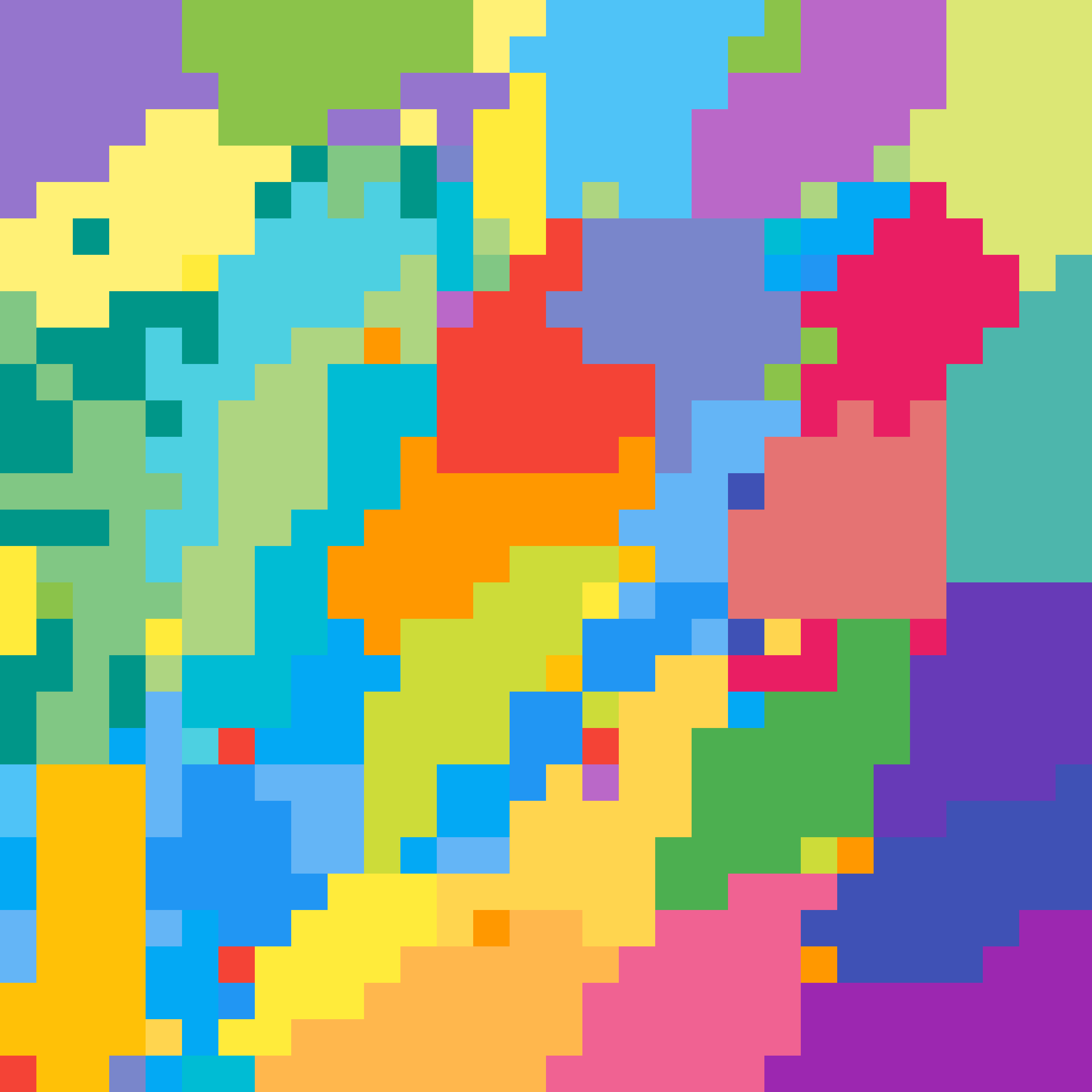}
	\caption{
		The same t-SNE found in Fig.~\ref{fig:tsne_img_correct} but using a unique color to denote the 3D object found in each rendered image. Here, each color also corresponds to a unique Inception-v3 label.
		Compare and contrast this plot with the t-SNE of only misclassified poses 
		(Fig.~\ref{fig:tsne_color_ax}).
		The original, high-resolution figure is available at \url{https://goo.gl/TGgPgB}.
	}\label{fig:tsne_color_correct}

\end{figure*}

\begin{figure*}
	\centering
	\includegraphics[width=1.0\columnwidth]{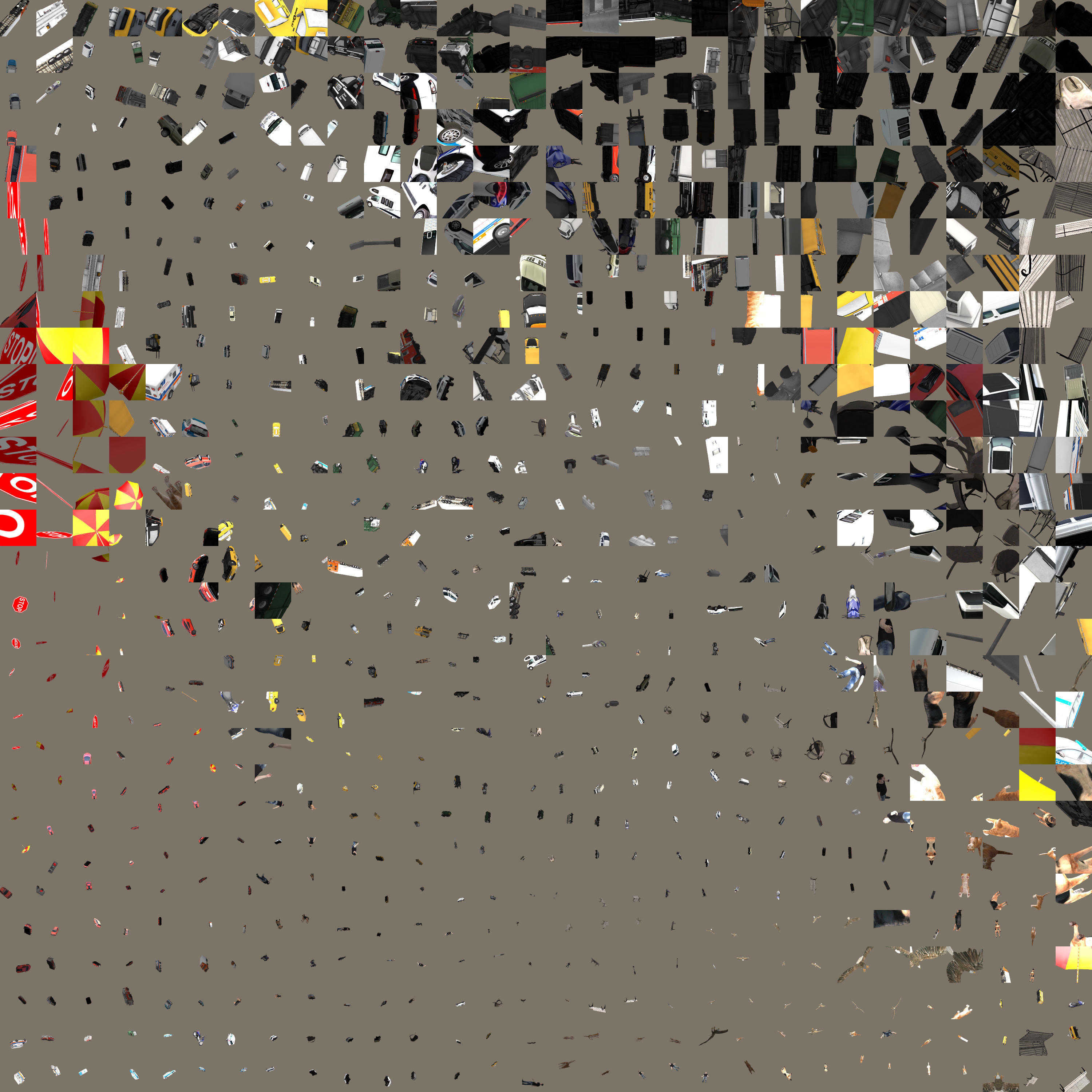}
	\caption{
		Following the same process as described in Fig.~\ref{fig:tsne_img_correct}, we show here a grid t-SNE of generated adversarial poses.
		For each object, we assembled 30 high-confidence ($p \geq 0.9$) adversarial examples generated via a random search against Inception-v3 \cite{szegedy2016rethinking}.
		The t-SNE was generated from the AlexNet \cite{Krizhevsky2012} \layer{fc7} features for $30$ objects $\times$ $30$ images = $900$ images.
		The original, high-resolution figure is available at \url{https://goo.gl/TGgPgB}.
		Adversarial poses were found to be both common across different objects (\eg, the top-right corner) and unique to specific objects (\eg, the \class{traffic sign} and \class{umbrella} objects in the middle left).
		\\To better understand how similar misclassified poses can be found across many objects, see Fig.~\ref{fig:tsne_color_ax}.
		Compare and contrast this plot with the t-SNE of correctly classified poses 
		(Figs.~\ref{fig:tsne_img_correct} \&~\ref{fig:tsne_color_correct}).
		%
%
	}\label{fig:tsne_img_ax}
\end{figure*}

\begin{figure*}
	\centering
	\includegraphics[width=1.0\columnwidth]{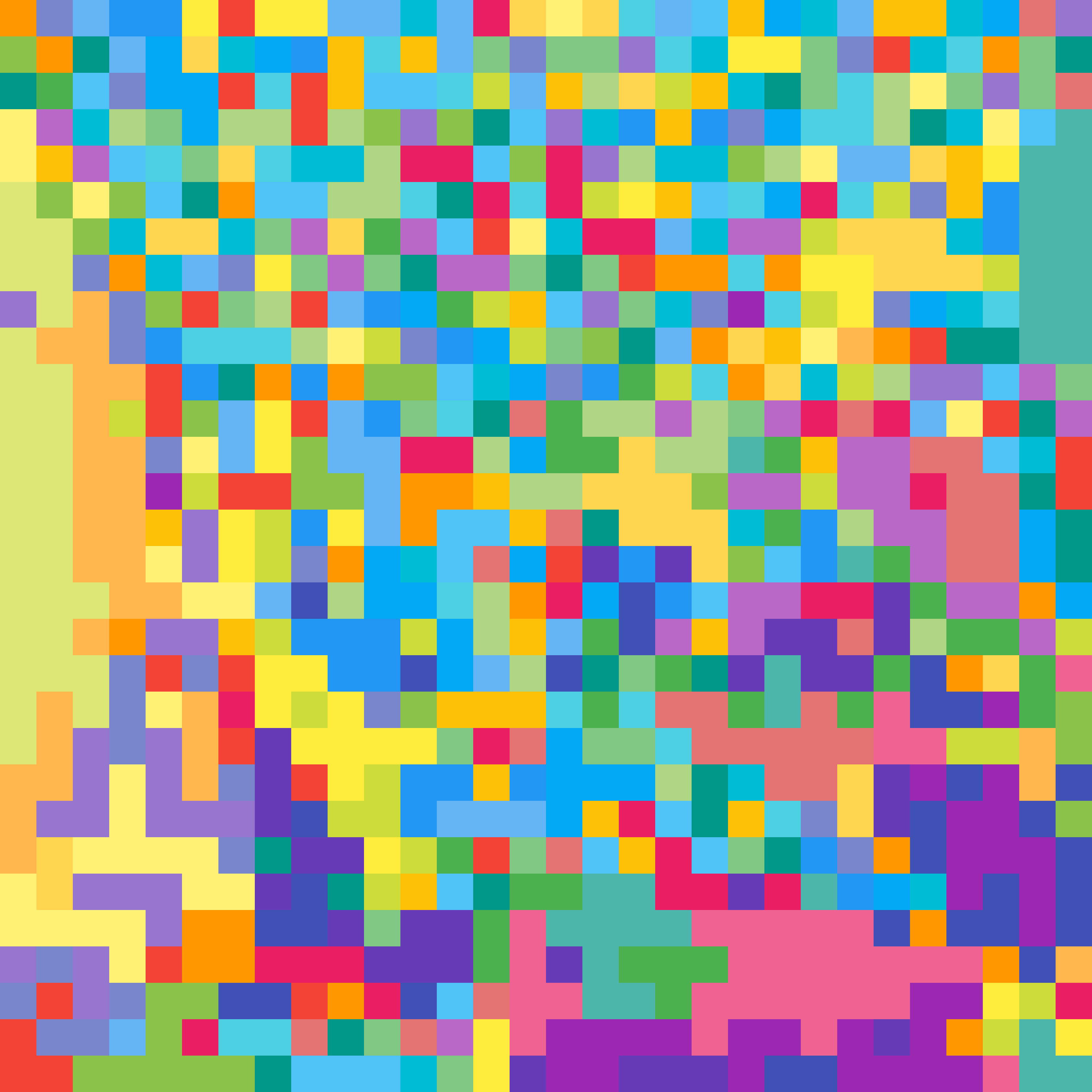}
	\caption{
		The same t-SNE as that in Fig.~\ref{fig:tsne_img_ax} but using a unique color to denote the 3D object used to render the adversarial image (\ie, Inception-v3's misclassification labels are not shown here). 
		The original, high-resolution figure is available at \url{https://goo.gl/TGgPgB}.\\
		Compare and contrast this plot with the t-SNE of correctly classified poses 
		(Fig.~\ref{fig:tsne_color_correct}).
		%
%
	}\label{fig:tsne_color_ax}
\end{figure*}

\section{Transferability from the Inception-v3 classifier to the YOLO-v3 detector}
\label{sec:yolo}

Previous research has shown that object detectors can be more robust to adversarial attacks than image classifiers~\cite{lu2017-standard}.
Here, we investigate how well our AXs generated for an Inception-v3 classifier trained to perform 1,000-way image classification on ImageNet~\cite{russakovsky2015imagenet} transfer to YOLO-v3, a state-of-the-art object detector trained on MS COCO~\cite{lin2014microsoft}.



Note that while ImageNet has 1,000 classes, MS COCO has bounding boxes classified into only 80 classes. 
Therefore, among 30 objects, we only selected the 13 objects that (1) belong to classes found in both the ImageNet and MS COCO datasets; and (2) are also well recognized by the YOLO-v3 detector in common poses.

\subsection{Class mappings from ImageNet to MS COCO}
See Table~\ref{tab:yolo_transfer_stats}a for 13 mappings from ImageNet labels to MS COCO labels.

\subsection{Selecting 13 objects for the transferability test}
For the transferability test (Sec.~\ref{sec:transferability}), we identified the 13 objects (out of 30) that are well detected by the YOLO-v3 detector via the two tests described below.

\subsubsection{YOLO-v3 correctly classifies 93.80\% of poses generated via yaw-rotation}
We rendered 36 unique views for each object by generating a render at every $30^\circ{}$ yaw-rotation (see Sec.~\ref{sec:quantitative_eval}).
Note that, across all objects, these yaw-rotation views have an average accuracy of $83.2\%$ by the Inception-v3 classifier.
We tested them against YOLO-v3 to see whether the detector was able to correctly find one single object per image and label it correctly.
Among 30 objects, we removed those that YOLO-v3 had an accuracy $\leq 70\%$, leaving 13 for the transferability test.
Across the remaining 13 objects, YOLO-v3 has an accuracy of 93.80\% on average (with an NMS threshold of $0.4$ and a confidence threshold of $0.5$).
Note that the accuracy was computed as the total number of correct labels over the total number of bounding boxes detected (\ie, we did not measure bounding-box IoU errors).
See class-specific statistics in Table~\ref{tab:yolo_transfer_stats}.
This result shows that YOLO-v3 is substantially more accurate than Inception-v3 on the standard object poses generated by yaw-rotation (93.80\% vs. 83.2\%).

\subsubsection{YOLO-v3 correctly classifies 81.03\% of poses correctly classified by Inception-v3}

Additionally, as a sanity check, we tested whether poses \emph{correctly classified} by Inception-v3 transfer well to YOLO-v3.
For each object, we randomly selected 30 poses that were $100\%$ correctly classified by Inception-v3 with high confidence ($p \geq 0.9$).
The images were generated via the random search procedure in the main text experiment (Sec.~\ref{sec:random_search}).
Across the final 13 objects, YOLO-v3 was able to correctly detect one single object per image and label it correctly at a 81.03\% accuracy (see Table~\ref{tab:yolo_transfer_stats}c).

\subsection{Transferability test: YOLO-v3 fails on 75.5\% of adversarial poses misclassified by Inception-v3}
\label{sec:transferability}

For each object, we collected 1,350 random adversarial poses (\ie, incorrectly classified by Inception-v3) generated via the random search procedure (Sec.~\ref{sec:random_search}).
Across all 13 objects and all adversarial poses, YOLO-v3 obtained an accuracy of only $24.50\%$ (compared to $81.03\%$ when tested on images correctly classified by Inception-v3).
In other words, 75.5\% of adversarial poses generated for Inception-v3 also escaped the detection\footnote{We were not able to check how many misclassification labels by YOLO-v3 were the same as those by Inception-v3 because only a small set of 80 the MS COCO classes overlap with the 1,000 ImageNet classes.} of YOLO-v3 (see Table~\ref{tab:yolo_transfer_stats}d for class-specific statistics).
Our result shows adversarial poses transfer well across tasks (image classification $\to$ object detection), models (Inception-v3 $\to$ YOLO-v3), and datasets (ImageNet $\to$ MS COCO).

\begin{table*}
	\centering
	\begin{tabular}{rllrrrrrrr}
		\toprule
		& \multicolumn{2}{c}{(a) Label mapping} & \multicolumn{2}{c}{(b) Accuracy on} & \multicolumn{2}{c}{(c) Accuracy on} & \multicolumn{3}{c}{(d) Accuracy on} \\
		& \multicolumn{2}{c}{} & \multicolumn{2}{c}{yaw-rotation poses} & \multicolumn{2}{c}{random poses} & \multicolumn{3}{c}{adversarial poses} \\
		\cmidrule(lr){2-3}\cmidrule(lr){4-5}\cmidrule(lr){6-7}\cmidrule(lr){8-10}
		& ImageNet & MS COCO & \#/36 & acc (\%) & \#/30 & acc (\%) & \#/1350 & acc (\%) & $\Delta$acc (\%) \\
		\midrule
		1 & park bench & bench & 31 & 86.11 & 22 & 73.33 & 211 & 15.63 & 57.70 \\
		2 & bald eagle & bird & 34 & 94.11 & 24 & 80.00 & 597 & 44.22 & 35.78 \\
		3 & school bus & bus & 36 & 100.00 & 18 & 60.00 & 4 & 0.30 & 69.70 \\
		4 & beach wagon & car & 34 & 94.44 & 30 & 100.00 & 232 & 17.19 & 82.81 \\
		5 & tiger cat & cat & 26 & 72.22 & 25 & 83.33 & 181 & 13.41 & 69.93 \\
		6 & German shepherd & dog & 32 & 88.89 & 28 & 93.33 & 406 & 30.07 & 63.26 \\
		7 & motor scooter & motorcycle & 36 & 100.00 & 18 & 60.00 & 384 & 28.44 & 31.56 \\
		8 & jean & person & 36 & 100.00 & 29 & 96.67 & 943 & 69.85 & 26.81 \\
		9 & street sign & stop sign & 31 & 86.11 & 26 & 86.67 & 338 & 25.04 & 61.15 \\
		10 & moving van & truck & 36 & 100.00 & 24 & 80.00 & 15 & 1.11 & 78.89 \\
		11 & umbrella & umbrella & 35 & 97.22 & 25 & 83.33 & 907 & 67.19 & 16.15 \\
		12 & police van & car & 36 & 100.00 & 25 & 83.33 & 55 & 4.07 & 79.26 \\
		13 & trailer truck & truck & 36 & 100.00 & 22 & 73.33 & 26 & 1.93 & 71.41 \\
		\midrule
		& &Average & & 93.80 & & 81.03 & &24.50 & 56.53 \\ \bottomrule
	\end{tabular}
	\caption{
		Adversarial poses generated for a state-of-the-art ImageNet image classifier (here, Inception-v3) transfer well to an MS COCO detector (here, YOLO-v3).
		The table shows the YOLO-v3 detector's accuracy on: (b) object poses generated by a standard process of yaw-rotating the object; (c) random poses that are $100\%$ correctly classified by Inception-v3 with high confidence ($p \geq 0.9$); and (d) adversarial poses, \ie, $100\%$ misclassified by Inception-v3.\\\\
		(a) The mappings of 13 ImageNet classes onto 12 MS COCO classes.\\
		(b) The accuracy (``acc (\%)'') of the YOLO-v3 detector on 36 yaw-rotation poses per object.\\
		(c) The accuracy of YOLO-v3 on 30 random poses per object that were correctly classified by Inception-v3.\\
		(d) The accuracy of YOLO-v3 on 1,350 adversarial poses (``acc (\%)'') and the differences between c and d (``$\Delta$acc (\%)'').
		%
	}
	\label{tab:yolo_transfer_stats}
\end{table*}

\section{Adversarial poses do exist in the real world}
\label{sec:transfer}

Our main experiments showed that adversarial poses exist in 3D simulation.
Here, we provide evidence that adversarial poses also transfer to and exist in the real world.

\begin{figure*}
	\begin{subfigure}{\linewidth}
		\centering
		\includegraphics[width=0.72\columnwidth]{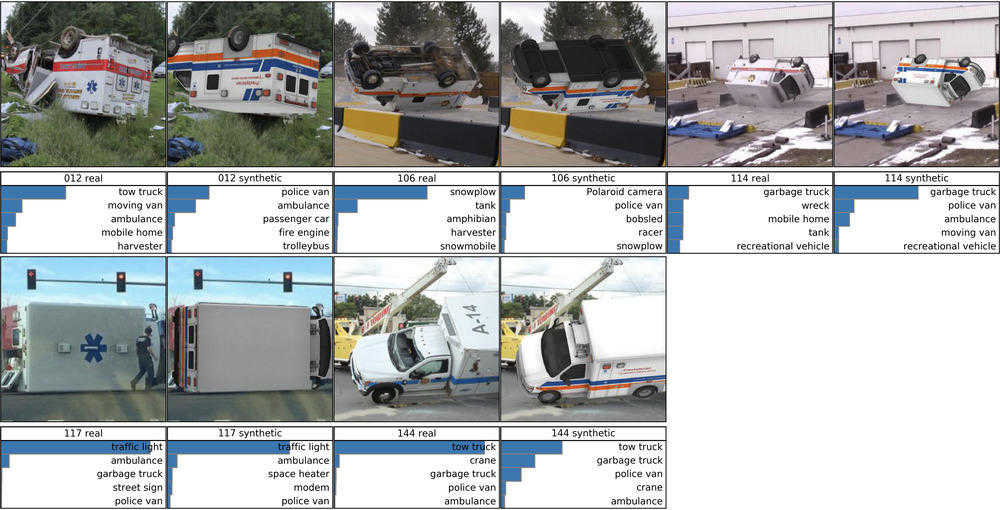}
		\caption{\class{ambulance}}\label{fig:transfer_ambulance}
	\end{subfigure}
	\begin{subfigure}{\linewidth}
		\centering
		\includegraphics[width=0.72\columnwidth]{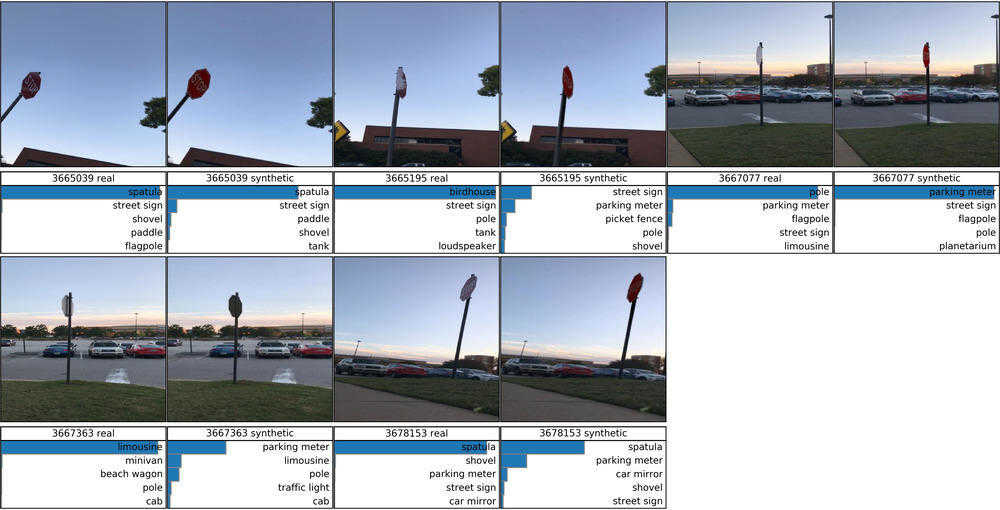}
		\caption{\class{street sign}}\label{fig:transfer_stop_sign}
	\end{subfigure}
	\begin{subfigure}{\linewidth}
		\centering
		\includegraphics[width=0.72\columnwidth]{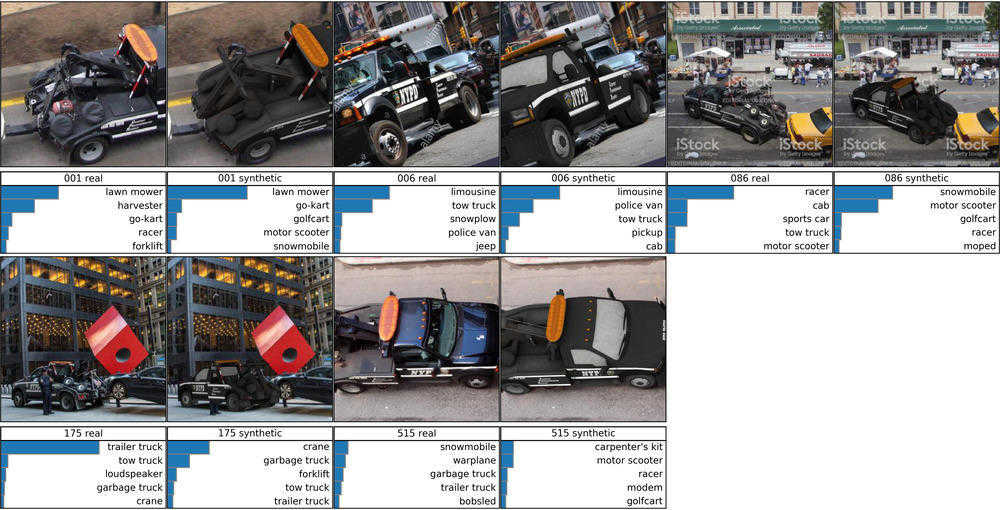}
		\caption{\class{tow truck}}\label{fig:transfer_tow_truck}
	\end{subfigure}
	\caption{
		Adversarial poses do transfer to the real world.
		We collected a set of 150 real photos ($5$ photos $\times$ $30$ objects) from the Internet that caused the Inception-v3 classifier to misclassify.
		For each pair, given the real, misclassified photo (left), we produced a render of the corresponding object (right) and gathered its top-5 predictions.
		We found that when the real photos appear out-of-distribution, $98.3\%$ of the renders are also misclassified, sometimes with the same top-1 label \eg, \class{spatula} in (b) or \class{lawn mower} in (c).
		Here, we show 5 pairs for each of the three example objects: (a) \class{ambulance}, (b) \class{street sign}, and (c) \class{tow truck}.
		The original, high-resolution figures for all 30 objects are available at \url{https://drive.google.com/open?id=18p-S9qO4dhE9toJbRRlAIWRcVqVH6Zsd}.}
		\label{fig:transfer}
\end{figure*}

First, we collected 5 photos $\times$ 30 objects $= 150$ photos from the Internet that were \emph{misclassified} by the Inception-v3 classifier and repeated the same experiment as described in Sec.~\ref{sec:qualitative_eval} to produce (real photo, render) pairs (see Fig.~\ref{fig:transfer}).
We found that when the real photos appear out-of-distribution, 98.3\% of the renders are also misclassified.
However, when the real failure photos appear ImageNet-like, $\sim$45\% of the renders are correctly classified (\ie, our 3D objects are \emph{easier} to recognize than their real-world counterparts).
This transferability result confirms the high realism of our 3D object renders and suggests that the adversarial poses do exist in the real world.

\begin{figure*}
    \begin{subfigure}{\linewidth}
        \centering
        \includegraphics[width=0.9\columnwidth]{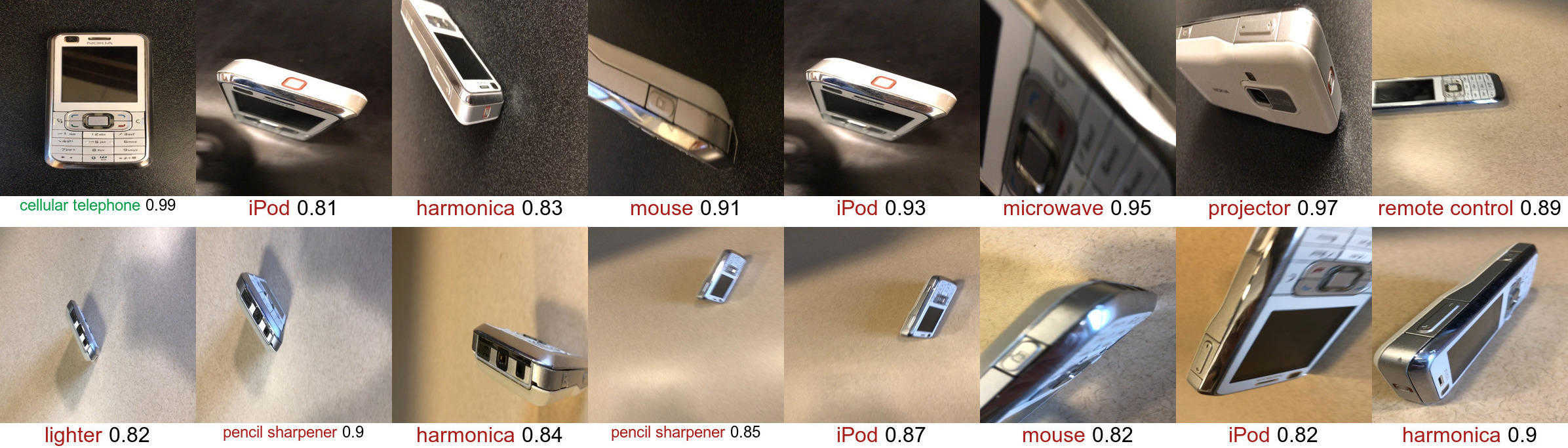}
        \caption{\class{cellular phone}}\label{fig:real_ax_cellphone}
    \end{subfigure}
    \begin{subfigure}{\linewidth}
        \centering
        \includegraphics[width=0.9\columnwidth]{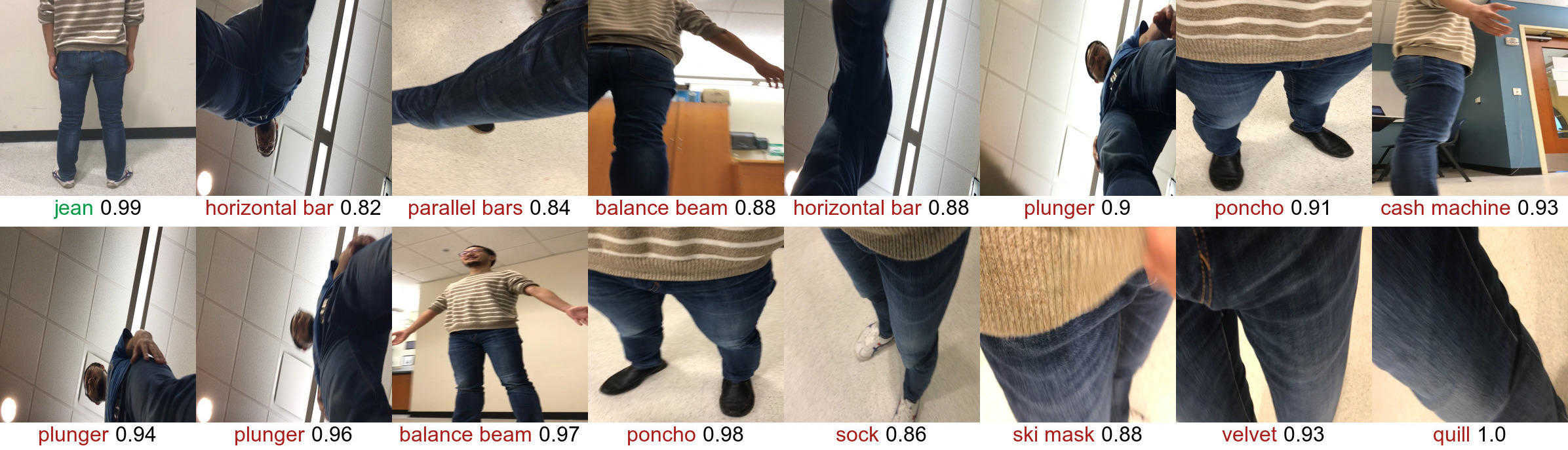}
        \caption{\class{jeans}}\label{fig:real_ax_jeans}
    \end{subfigure}
    \begin{subfigure}{\linewidth}
        \centering
        \includegraphics[width=0.9\columnwidth]{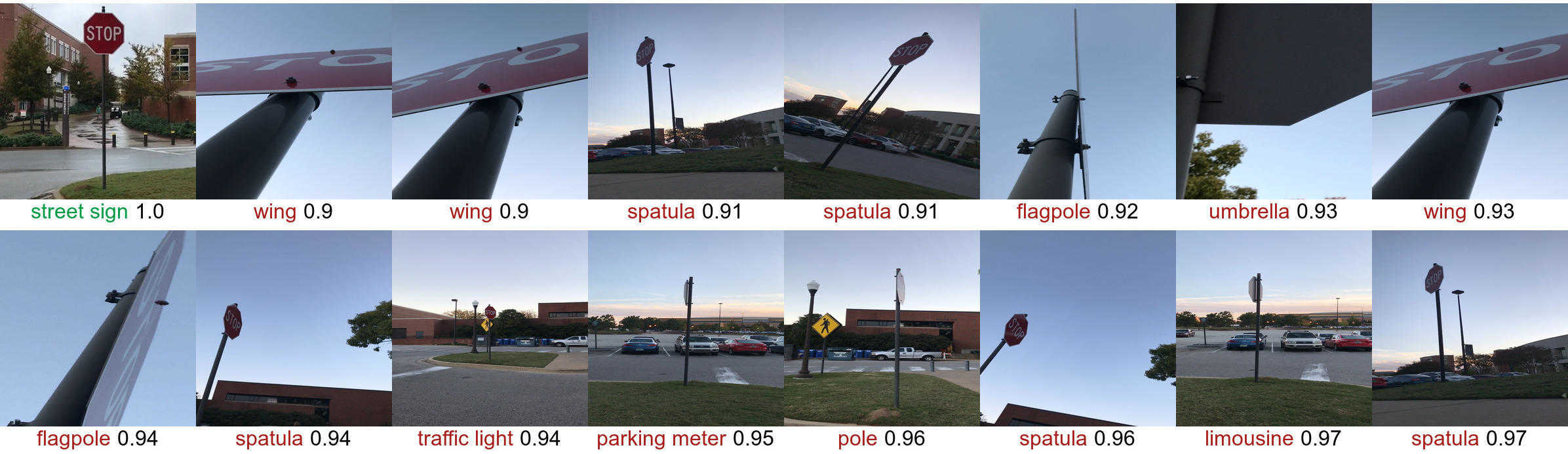}
        \caption{\class{street sign}}\label{fig:real_ax_stopsign}
    \end{subfigure}
    \begin{subfigure}{\linewidth}
        \centering
        \includegraphics[width=0.9\columnwidth]{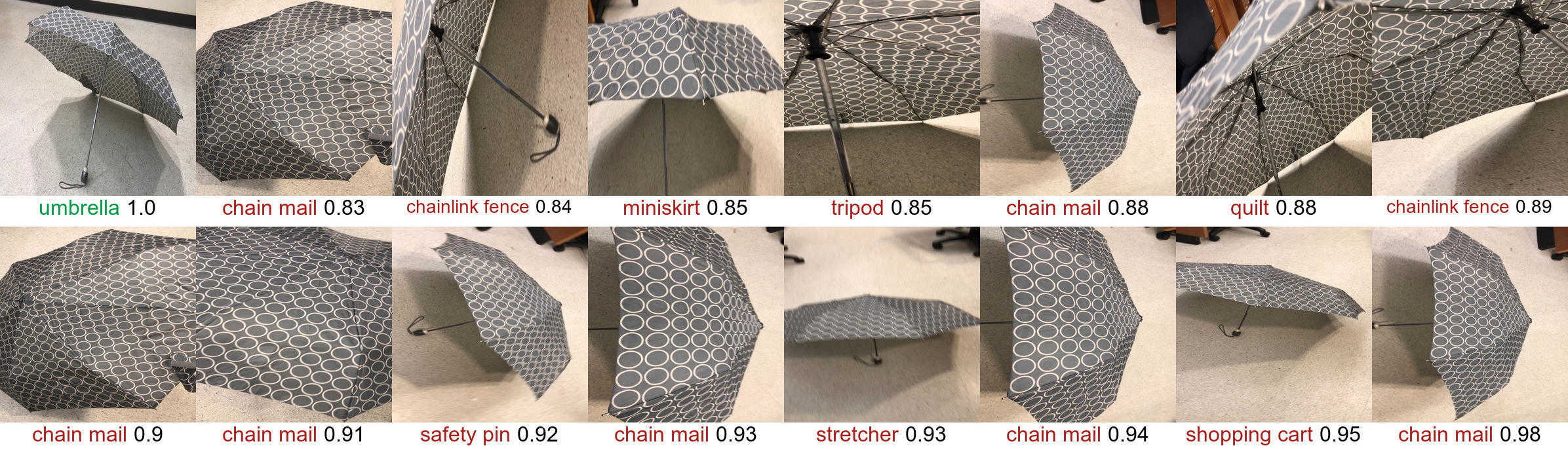}
        \caption{\class{umbrella}}\label{fig:real_ax_umbrella}
    \end{subfigure}
\caption{
     Real-world, high-confidence adversarial poses can be found by taking photos from strange angles of a familiar object, here, \class{cellular phone}, \class{jeans}, \class{street sign}, and \class{umbrella}.
     While Inception-v3 \cite{szegedy2016rethinking} can correctly predict the object in canonical poses (the top-left image in each panel), the model misclassified the same objects in unusual poses.
     Below each image is its top-1 prediction label and confidence score.
     We took real-world videos of these four objects and extracted these misclassified poses from the videos.
	The original, high-resolution figures are available at \url{https://goo.gl/zDWcjG}.
}\label{fig:real_ax}
\end{figure*}

Second, we found that real-world, high-confidence adversarial poses can be found by simply taking photos from strange angles of a familiar object.
We took real-world videos of four example objects (\class{cellular phone}, \class{jeans}, \class{street sign}, and \class{umbrella}) and extracted the misclassified frames from the videos.
While Inception-v3 \cite{szegedy2016rethinking} correctly recognized these objects in canonical poses, the model misclassified the same objects in unusual poses (Fig.~\ref{fig:real_ax}).

\clearpage

\section{Experimental setup for the differentiable renderer}\label{sec:dr-expr}

For the gradient descent method (DR-G) that uses the approximate gradients provided by the differentiable renderer \cite{kato2018neural} (DR), we set up the rendering parameters in the DR to closely match those in the NR.
However, there were still subtle discrepancies between the DR and the NR that made the results (DR-G vs. FD-G in Sec.~\ref{sec:comparing_methods}) not directly comparable.
Despite these discrepancies (described below), we still believe the FD gradients are more stable and informative than the DR gradients (\ie, FD-G outperformed DR-G).\footnote{In preliminary experiments with only the DR (not the NR), we also empirically found FD-G to be more stable and effective than DR-G (data not shown).}\\

\begin{figure}[ht]
  \centering
  \subcaptionbox{School bus at \((0, 0, -14)\)}{{\includegraphics[width=.32\textwidth]{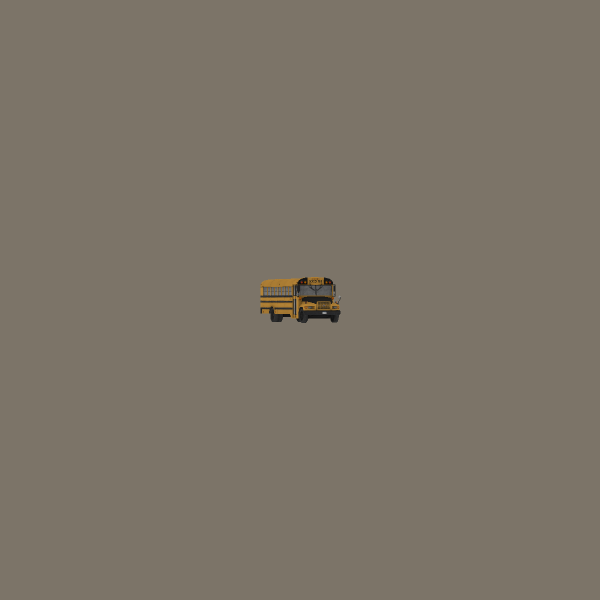}}}
  \subcaptionbox{School bus at \((0, 0, 0)\)}{{\includegraphics[width=.32\textwidth]{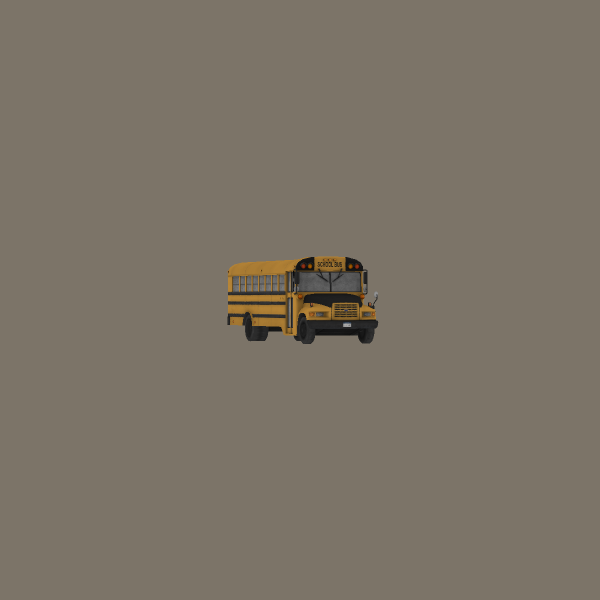}}}
  \subcaptionbox{School bus at \((0, 0, 14)\)}{{\includegraphics[width=.32\textwidth]{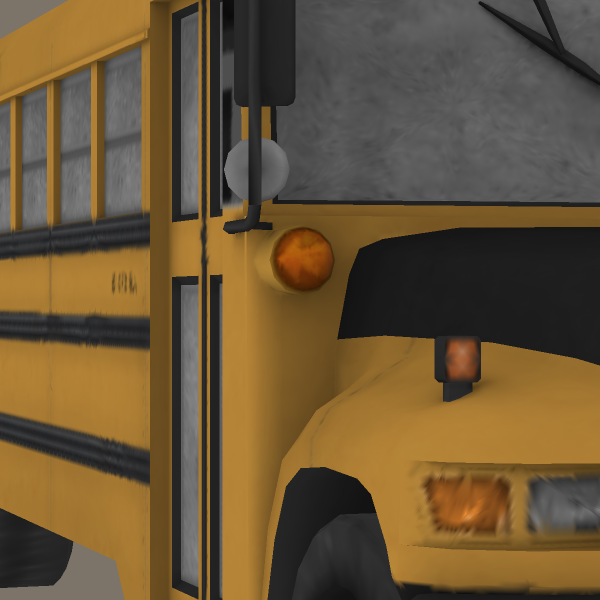}}}
  \caption{School bus rendered by the DR at different distances.}\label{fig:dr-demo}
\end{figure}

\subsec{DR setup} 
For all experiments with the DR, the camera was centered at \((0, 0, 16)\) with an up direction \((0, 1, 0)\). 
The object's spatial location was constrained such that the object center was always within the frame.
The depth values were constrained to be within \([-14, 14]\).
Similar to experiments with the NR, we used the \medium lighting setting.
The ambient light color was set to white with an intensity 1.0, while the directional light was set to white with an intensity 0.4.  
Fig.~\ref{fig:dr-demo} shows an example school bus rendered under this \medium lighting at different distances.

The known discrepancies between the experimental setups of FD-G (with the NR) vs. DR-G (with the DR) are:

\begin{enumerate}
	\item The exact \medium lighting parameters for the NR described in the main text (Sec.~\ref{sec:easily_confused}) did not produce similar lighting effects in the DR.
	Therefore, the DR lighting parameters described above were the result of manually tuning to qualitatively match the effect produced by the NR \medium lighting parameters.
	\item While the NR uses a built-in tessellation procedure that automatically tessellates input objects before rendering, we had to perform an extra pre-processing step of manually tessellating each object for the DR.
	While small, a discrepancy still exists between the two rendering results (Fig.~\ref{fig:compare_tessellation}b vs. c).
\end{enumerate}

\section{Gradient descent with the DR gradients}\label{sec:kr}

In preliminary experiments (data not shown), we found the DR gradients to be relatively noisy when using gradient descent to find targeted adversarial poses (\ie, DR-G experiments).
To mitigate this problem, we experimented with (1) parameter augmentation (Sec.~\ref{sec:param-augm}); and (2) multi-view optimization (Sec.~\ref{sec:multi-camera-optim}).
In short, we found parameter augmentation helped and used it in DR-G.
However, when using the DR, we did not find multiple cameras improved optimization performance and thus only performed regular single-view optimization for DR-G.

\subsection{Parameter augmentation}%
\label{sec:param-augm}

We performed gradient descent using the DR gradients (DR-G) in an augmented parameter space corresponding to 50 rotations and one translation to be applied to the original object vertices.
That is, we backpropagated the DR gradients into the parameters of these pre-defined transformation matrices.
Note that DR-G is given the same budget of $100$ steps per optimization run as FD-G and ZRS for comparison in Sec.~\ref{sec:comparing_methods}.

The final transformation matrix is constructed by a series of rotations followed
by one translation, i.e., 

\begin{align*}
M = T\cdot R_{n-1}R_{n-2}\cdots R_0
\end{align*}

\noindent where \(M\) is the final transformation matrix, \(R_i\) the rotation matrices, and \(T\) the translation matrix.  

\begin{figure}[h]
  \centering
  \subcaptionbox{$y$-axis: success rate}{\includegraphics[width=0.25\linewidth]{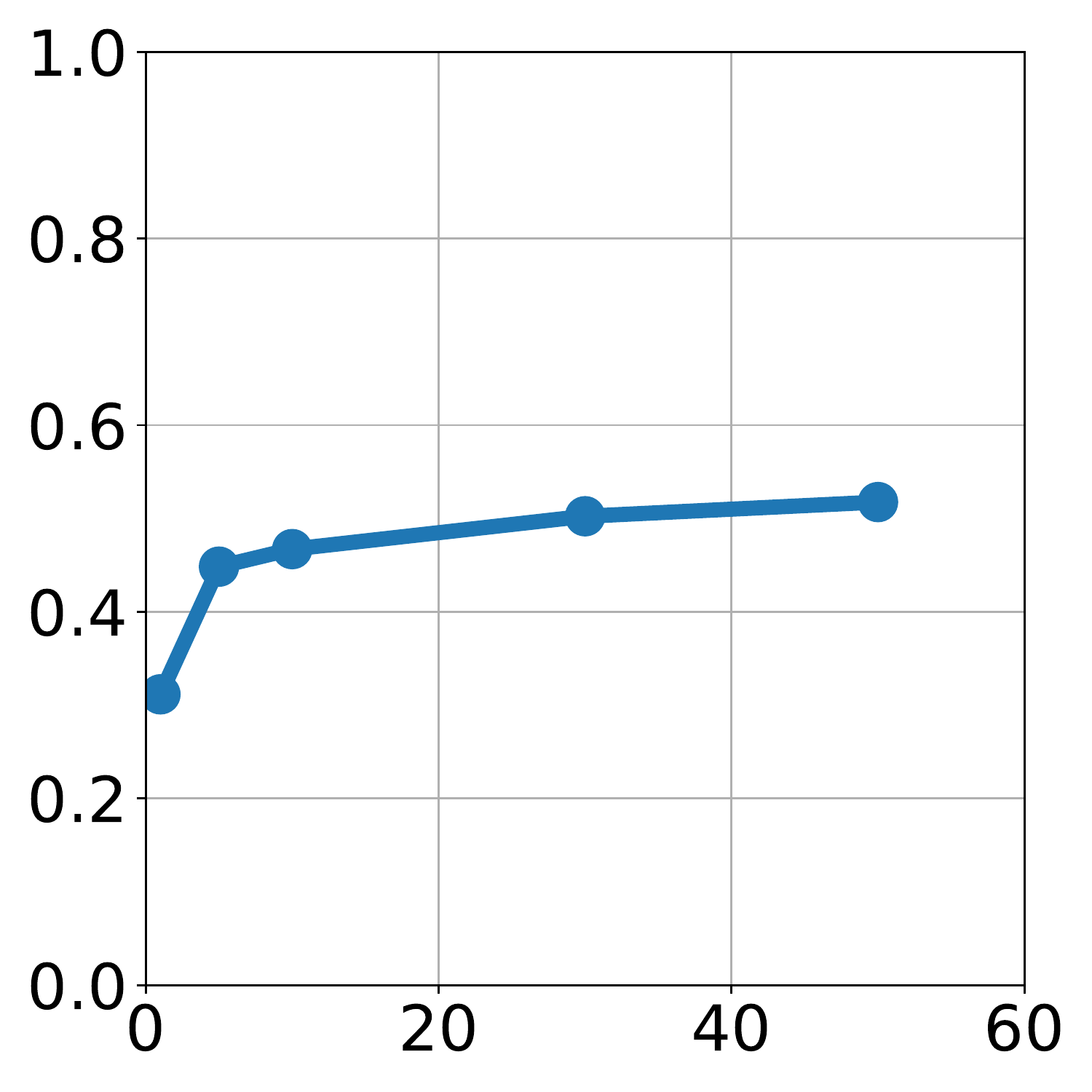}}%
  \subcaptionbox{$y$-axis: max confidence}{\includegraphics[width=0.25\linewidth]{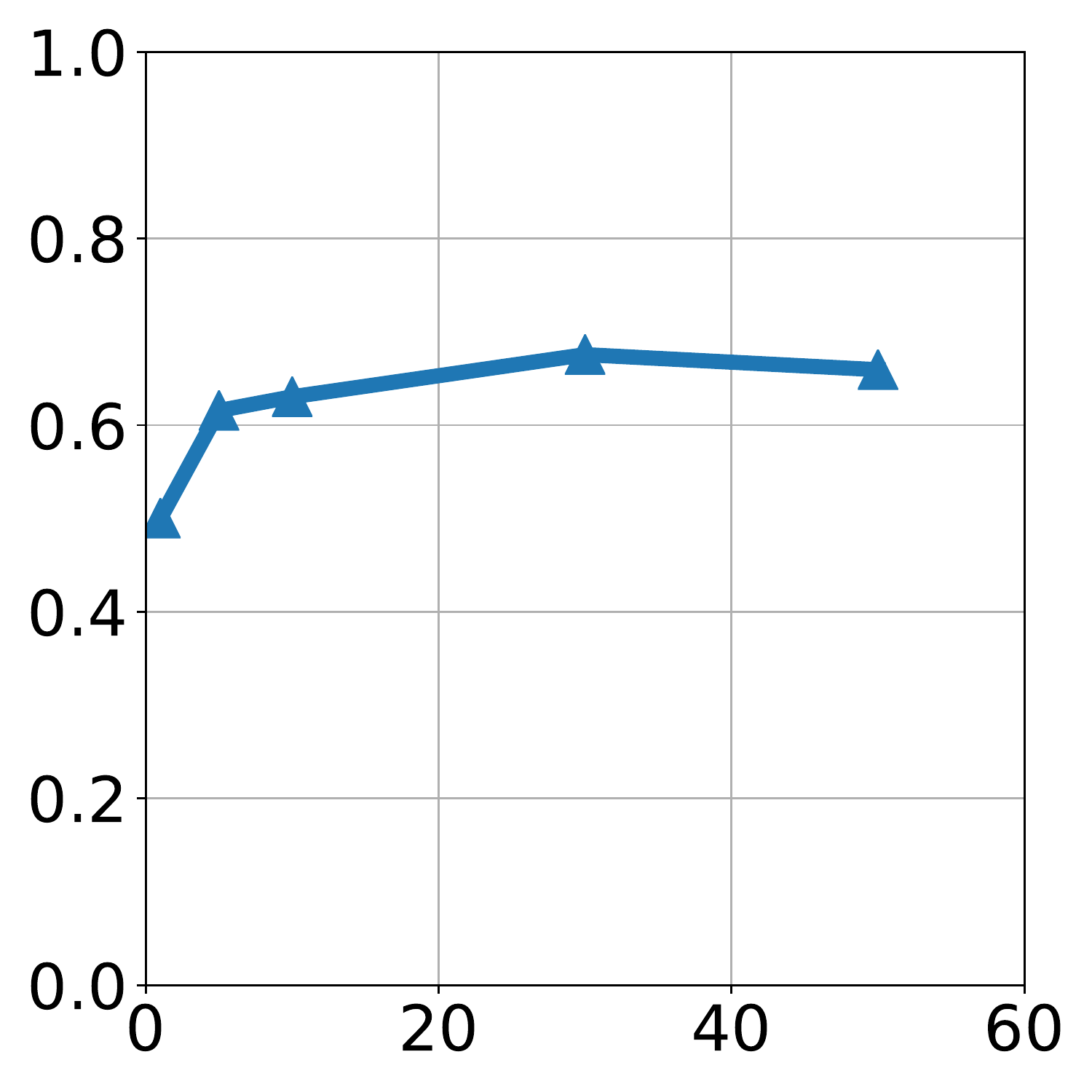}}%
  \subcaptionbox{$y$-axis: mean number of steps}{\includegraphics[width=0.25\linewidth]{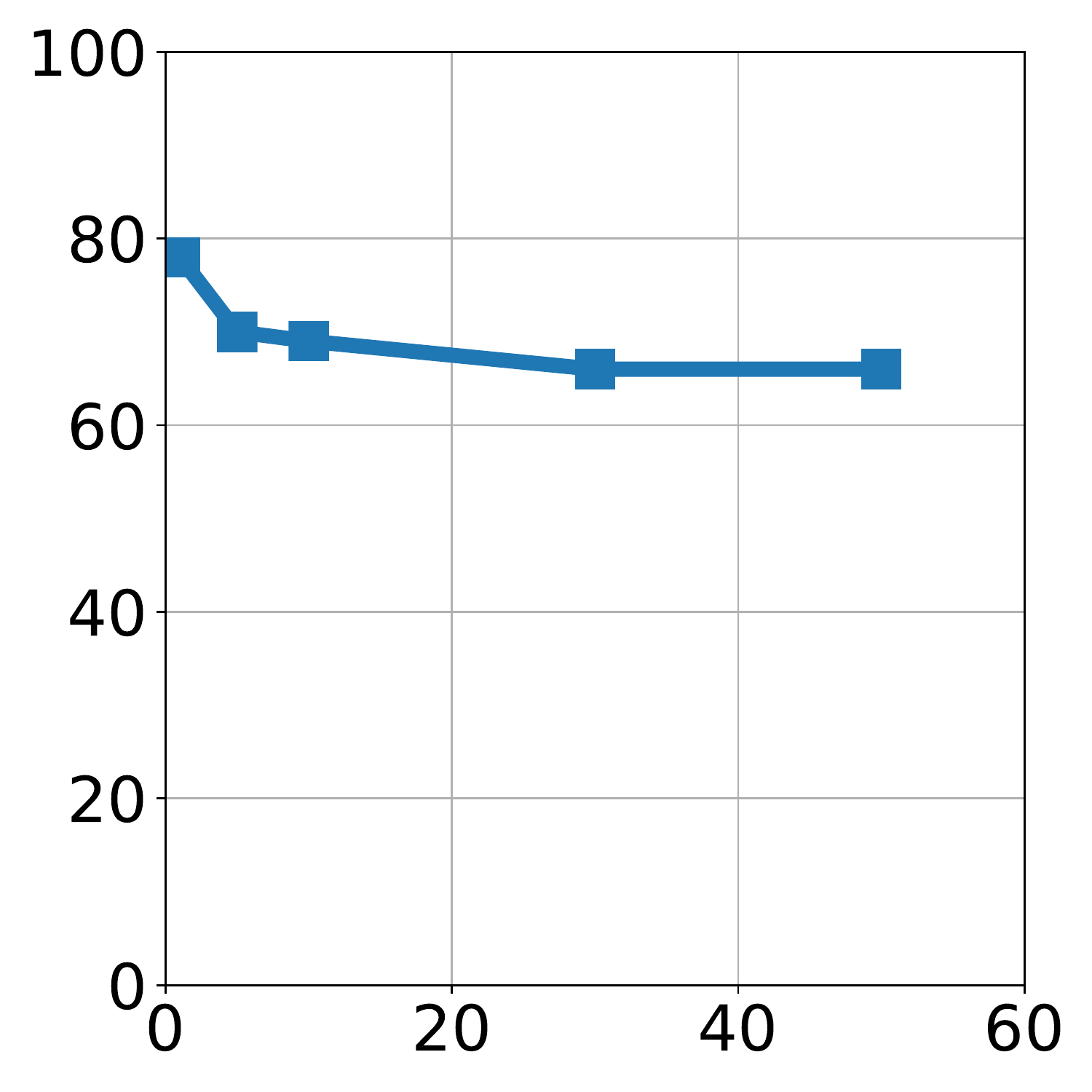}}
	\caption{
  	We found that increasing the number of rotations (displayed in $x$-axes) per step helped:\\
  	(a) improve the success rate of hitting the target labels;\\
  	(b) increase the maximum confidence score of the found adversarial examples;\\
  	(c) reduce the average number of steps required to find an AX, \ie, led to faster convergence.\\
}
\label{fig:param-augm}
\end{figure}

We empirically found that increasing the number of rotations per step helped (a) improve the success rate of hitting the target labels; (b) increase the maximum confidence score of the found AXs; and (c) reduce the number of steps, \ie, led to faster convergence (see Fig.~\ref{fig:param-augm}).
Therefore, we empirically chose $n=50$ for all DR-G experiments reported in the main text.


\subsection{Multi-view optimization}%
\label{sec:multi-camera-optim}

Additionally, we attempted to harness multiple views (from multiple cameras) to increase the chance of finding a target adversarial pose.
Multi-view optimization did not outperform single-view optimization using the DR in our experiments.
Therefore, we only performed regular single-view optimization for DR-G.
We briefly document our negative results below.

Instead of backpropagating the DR gradient to a single camera looking at the object in the 3D scene, one may set up multiple cameras, each looking at the object from a different angle.
This strategy intuitively allows gradients to still be backpropagated into the vertices that may be occluded in one view but visible in some other view.
We experimented with six cameras and backpropagating to all cameras in each step.
However, we only updated the object following the gradient from the view that yielded the lowest loss among all views.
One hypothesis is that having multiple cameras might improve the chance of hitting the target.

In our experiments with the DR using 100 steps per optimization run, multi-view optimization performed worse than single-view in terms of both the success rate and the number of steps to converge.
We did not compare all 30 objects due to the expensive computational cost, and only report the results from optimizing two objects \class{bald eagle} and \class{tiger cat} in Table~\ref{tab:v1v6}.
Intuitively, multi-view optimization might outperform single-view optimization given a large enough number of steps.

\begin{table}[ht]
  \centering
  \begin{tabular}{l*{4}{r}}
    \toprule
    & \multicolumn{2}{c}{
    	\class{bald eagle}} & \multicolumn{2}{c}{\class{tiger cat}} \\
    \cmidrule(lr){2-3}\cmidrule(lr){4-5}
    & Steps & Success rate & Steps & Success rate \\
    \midrule
    Single-view  & 71.80 & 0.44 & 90.70 & 0.15 \\
    Multi-view & 81.28 & 0.23 & 96.84 & 0.04 \\
    \bottomrule
  \end{tabular}
  \caption{
  	Multi-view optimization performed worse than single-view optimization in both (a) the number of steps to converge and (b) success rates.
  	We show here the results of two runs of optimizing with the \class{bald eagle} and \class{tiger cat} objects.
  	The results are averaged over $50$ target labels $\times 50$ trials = $2,500$ trials.
  	Each optimization trial for both single- and multi-view settings is given the budget of $100$ steps.
  	\label{tab:v1v6}}
\end{table}

\newpage
\section{3D transformation matrix}\label{sec:trans-mat}

A rotation of \(\theta\) around an arbitrary axis \((x, y, z)\) is given by the
following homogeneous transformation matrix.

\begin{equation}
  \label{eq:rotation-matrix}
  R =
  \begin{vmatrix*}[l]
    x x(1 - c) + c & x y(1 - c) - z s & x z(1 - c) + y s & 0\\
    x y(1 - c) + z s & y y(1 - c) + c & y z(1 - c) - x s & 0\\
    x z(1 - c) - y s & y z(1 - c) + x s & y z(1 - c) + c & 0\\
    0 & 0 & 0 & 1
  \end{vmatrix*}
\end{equation}

\noindent
where \(s = \sin\theta\), \(c = \cos\theta\), and the axis is normalized, \ie, \(x^2 + y^2 + z^2 = 1\).
Translation by a vector \((x, y, z)\) is given by the following homogeneous
transformation matrix.

\begin{equation}
  \label{eq:translation-matrix}
  T =
  \begin{vmatrix*}[l]
    1 & 0 & 0 & x\\
    0 & 1 & 0 & y\\
    0 & 0 & 1 & z\\
    0 & 0 & 0 & 1
  \end{vmatrix*}
\end{equation}

Note that in the optimization experiments with random search (RS) and finite-difference gradients (FD-G), we dropped the homogeneous component for simplicity, \ie, the rotation matrices of yaw, pitch, and roll are all $3 \times 3$.  
The homogeneous component is only necessary for translation, which can be achieved via simple vector addition. 
However, in DR-G, we used the homogeneous component because we had some experiments interweaving translation and rotation.  
The matrix representation was more convenient for the DR-G experiments.  
As they are mathematically equivalent, this arbitrary implementation choice should not alter our results.

\newpage

\begin{table*}[h]
	\begin{center}
		\begin{tabular}{lr}
			\toprule
			Object          & Accuracy (\%) \\
			\midrule
			ambulance       & 3.64       \\
			backpack        & 8.63       \\
			bald eagle      & 13.26      \\
			beach wagon     & 0.60       \\
			cab             & 2.64       \\
			cell phone      & 14.97      \\
			fire engine     & 4.31       \\
			forklift        & 5.20       \\
			garbage truck   & 4.88       \\
			German shepherd & 9.61       \\
			\bottomrule
		\end{tabular}
		\begin{tabular}{lr}
			\toprule
			Object        & Accuracy (\%) \\
			\midrule
			golfcart      & 2.14       \\
			jean          & 2.71       \\
			jeep          & 0.29       \\
			minibus       & 0.83       \\
			minivan       & 0.66       \\
			motor scooter & 20.49      \\
			moving van    & 0.45       \\
			park bench    & 5.72       \\
			parking meter & 1.27       \\
			pickup        & 0.86       \\
			\bottomrule
		\end{tabular}
		\begin{tabular}{lr}
			\toprule
			Object               & Accuracy (\%) \\
			\midrule
			police van           & 0.95       \\
			recreational vehicle & 2.05       \\
			school bus           & 3.48       \\
			sports car           & 2.50       \\
			street sign          & 26.32      \\
			tiger cat            & 7.36       \\
			tow truck            & 0.87       \\
			traffic light        & 14.95      \\
			trailer truck        & 1.27       \\
			umbrella             & 49.88      \\
			\bottomrule
		\end{tabular}
	\end{center}
	\caption{The percent of three million random samples
		that were correctly
		classified by Inception-v3 \cite{szegedy2016rethinking} for each object.
		That is, for each lighting setting in $\{ \bright, \medium, \dark\}$, we generated $10^6$ samples.
		See Sec.~\ref{sec:random_search}
		for details on the sampling procedure.
	}
	\label{tab:sampling_stats}
\end{table*}


\begin{figure}[ht]
	\centering
	\subcaptionbox{\bright}{{\includegraphics[width=.32\textwidth]{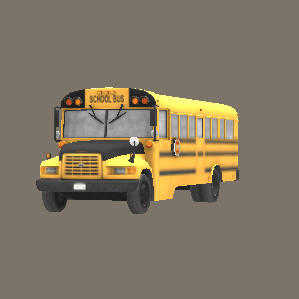}}}
	\subcaptionbox{\medium}{{\includegraphics[width=.32\textwidth]{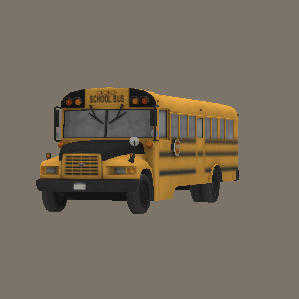}}}
	\subcaptionbox{\dark}{{\includegraphics[width=.32\textwidth]{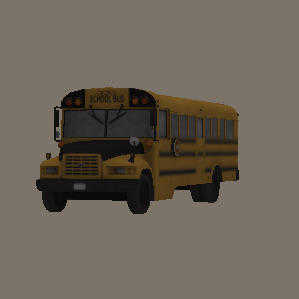}}}
	\caption{Renders of the \class{school bus} object using the NR \cite{modernGL} at three different lighting settings. 
	The directional light intensities and ambient light intensities were $(1.2, 1.6)$, $(0.4, 1.0)$, and $(0.2, 0.5)$ for the \bright, \medium, and \dark settings, respectively.}\label{fig:light_intensity}
\end{figure}

\section{Adversarial poses were not found in ImageNet classes via a nearest-neighbor search}
\label{sec:nearest_neighbors}

We performed a nearest-neighbor search to check whether adversarial poses generated (in Sec.~\ref{sec:easily_confused}) can be found in the ImageNet dataset.

\begin{figure*}
    \centering
    \includegraphics[width=0.9\columnwidth]{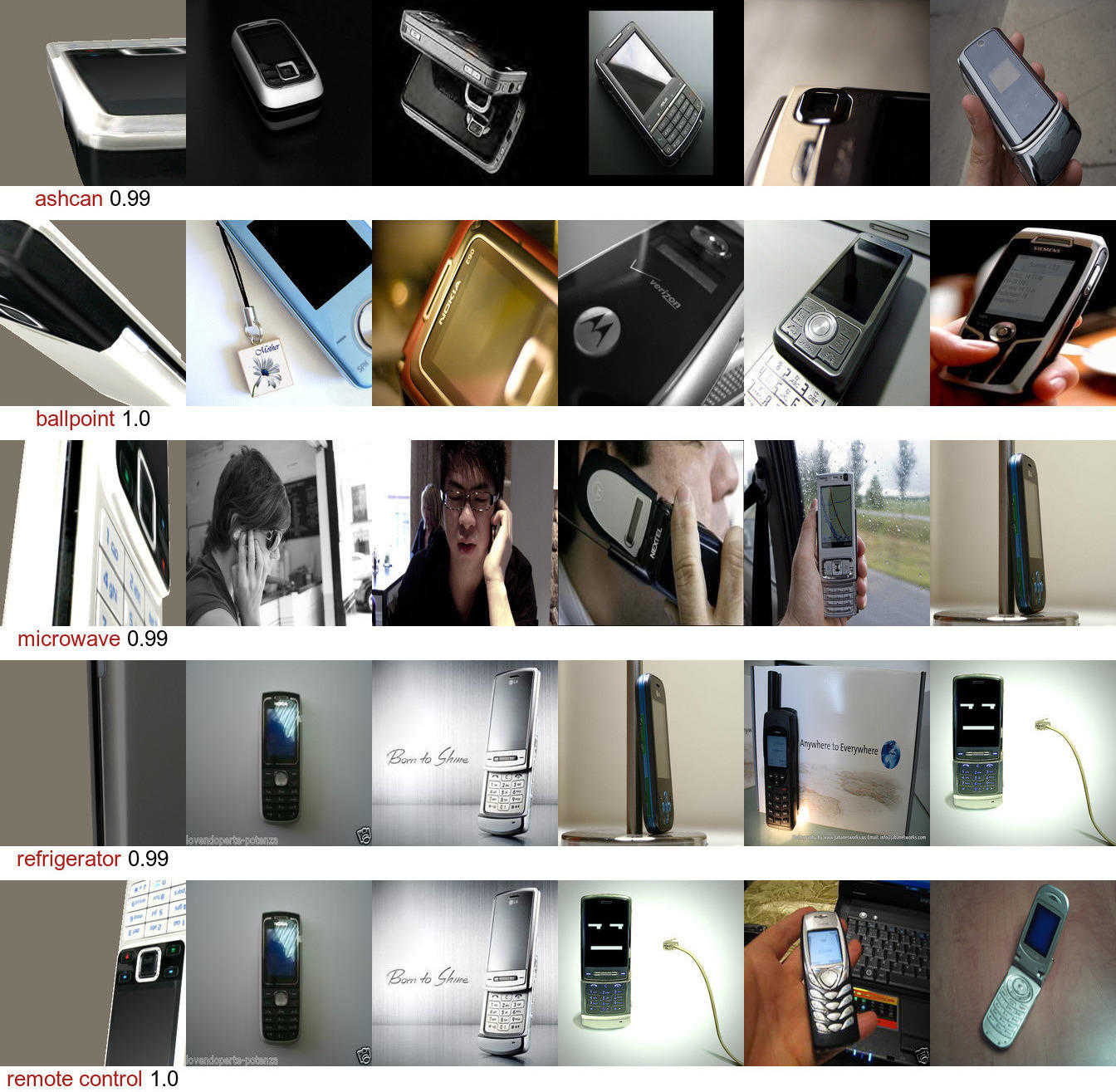}
    \caption{
    For each adversarial example (leftmost), we retrieved the five nearest neighbors (five rightmost photos) from all $\sim$1,300 images in the \class{cellular phone} class.
    The Euclidean distance between a pair of images was computed in the \layer{fc7} feature space of a pre-trained AlexNet \cite{Krizhevsky2012}.
   	The nearest photos from the class are mostly different from the adversarial poses.
    This result supports the hypothesis that the generated adversarial poses are out-of-distribution.
    The original, high-resolution figure is available at \url{https://goo.gl/X31VXh}.}
    \label{fig:nearest_cellphone}
\end{figure*}

\begin{figure*}
    \centering
    \includegraphics[width=0.9\columnwidth]{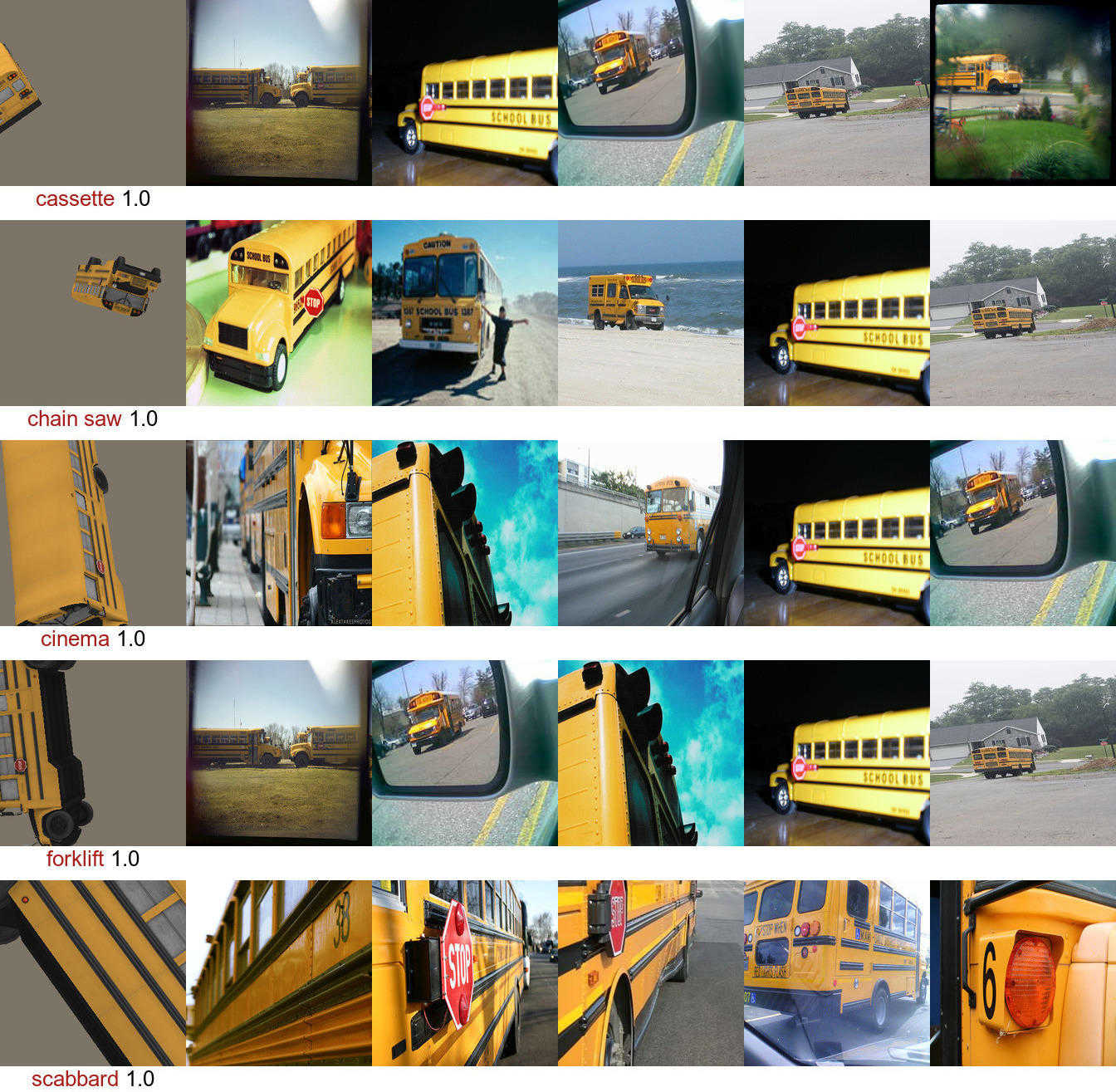}
    \caption{
	For each adversarial example (leftmost), we retrieved the five nearest neighbors (five rightmost photos) from all $\sim$1,300 images in the \class{school bus} class.
	The Euclidean distance between a pair of images was computed in the \layer{fc7} feature space of a pre-trained AlexNet \cite{Krizhevsky2012}.
	The nearest photos from the class are mostly different from the adversarial poses.
	This result supports the hypothesis that the generated adversarial poses are out-of-distribution.
	The original, high-resolution figure is available at \url{https://goo.gl/X31VXh}.}
    \label{fig:nearest_schoolbus}
\end{figure*}

\begin{figure*}
    \centering
    \includegraphics[width=0.9\columnwidth]{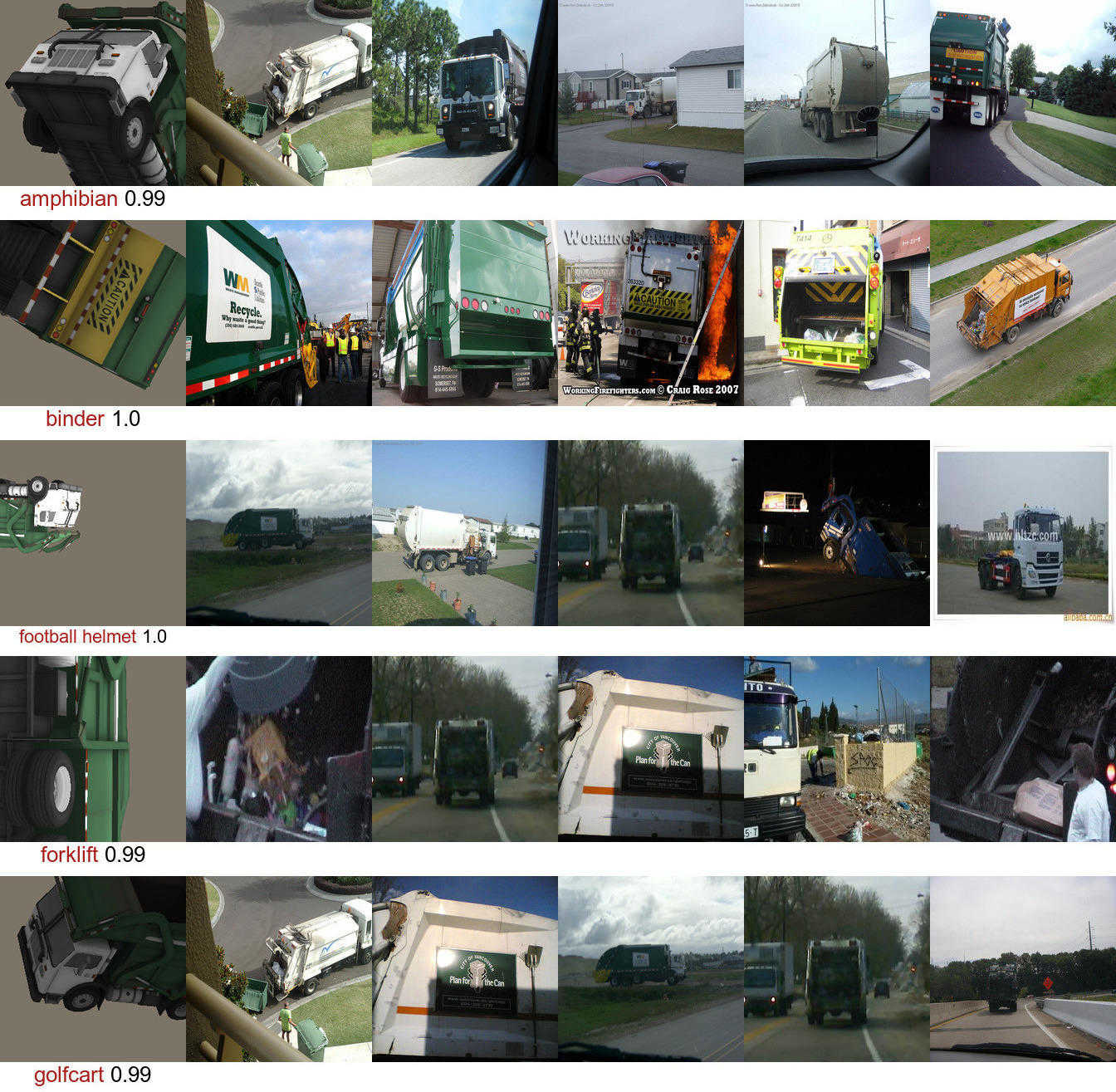}
    \caption{
    For each adversarial example (leftmost), we retrieved the five nearest neighbors (five rightmost photos) from all $\sim$1,300 images in the \class{garbage truck} class.
    The Euclidean distance between a pair of images was computed in the \layer{fc7} feature space of a pre-trained AlexNet \cite{Krizhevsky2012}.
	The nearest photos from the class are mostly different from the adversarial poses.
	This result supports the hypothesis that the generated adversarial poses are out-of-distribution.
    The original, high-resolution image is available at \url{https://goo.gl/X31VXh}.}
    \label{fig:nearest_garbagetruck}
\end{figure*}

~\\\subsec{Retrieving nearest neighbors from a single class corresponding to the 3D object} 
We retrieved the five nearest training-set images for each adversarial pose (taken from a random selection of adversarial poses) using the \layer{fc7} feature space from a pre-trained AlexNet \cite{Krizhevsky2012}.
The Euclidean distance was used to measure the distance between two \layer{fc7} feature vectors.
We did not find qualitatively similar images despite comparing all $\sim$1,300 class images corresponding to the 3D object used to generate the adversarial poses (\eg, \class{cellphone}, \class{school bus}, and \class{garbage truck} in Figs.~\ref{fig:nearest_cellphone},~\ref{fig:nearest_schoolbus}, and~\ref{fig:nearest_garbagetruck}).
This result supports the hypothesis that the generated adversarial poses are out-of-distribution.

\begin{figure*}[t]
	\begin{center}
		\includegraphics[width=1.0\textwidth]{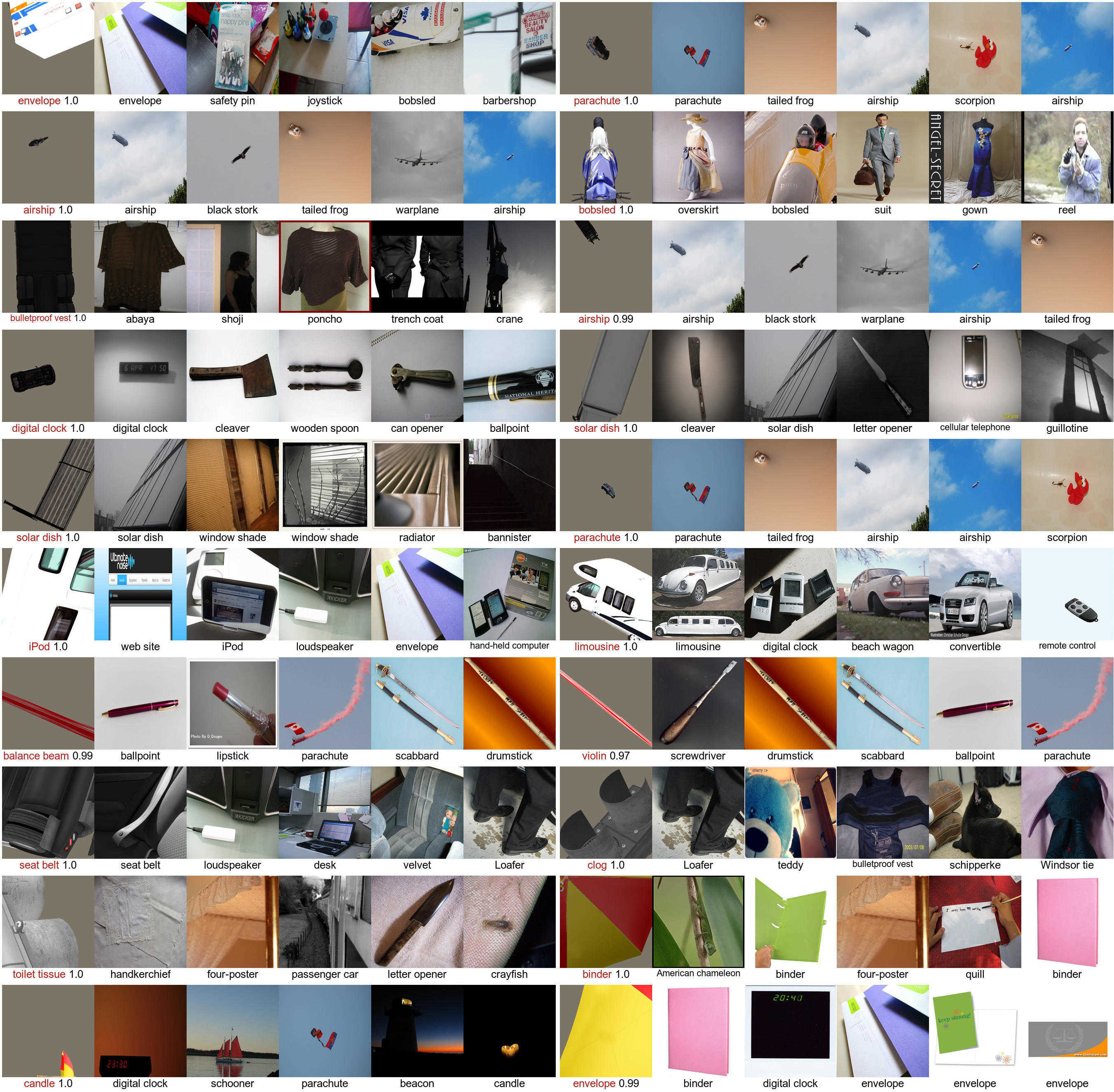}
	\end{center}
	\caption{
		For each adversarial example (leftmost), we retrieved the five nearest neighbors (five rightmost photos) from the 50,000-image ImageNet validation set.
		The Euclidean distance between a pair of images was computed in the \layer{fc7} feature space of a pre-trained AlexNet \cite{Krizhevsky2012}.
		Below each adversarial example (AX) is its Inception-v3 \cite{szegedy2016rethinking} top-1 prediction label and confidence score.
		The associated ground-truth ImageNet label is beneath each retrieved photo.
		Here, we show an interesting, cherry-picked collection of cases where the nearest photos (in the \layer{fc7} feature space) are also qualitatively similar to the reference AX and sometimes come from the exact same class as the AX's predicted label.
		More examples are available at \url{https://goo.gl/8ib2PR}.
		}
	\label{fig:nearest_val_images}
\end{figure*}

~\\\subsec{Searching from the validation set}
We also searched the entire 50,000-image validation set of ImageNet.
Interestingly, we found the top-5 nearest images were sometimes from the same class as the \emph{targeted} misclassification label (see Fig.~\ref{fig:nearest_val_images}).

%
%



\begin{figure*}
	\centering
	\includegraphics[width=0.76\columnwidth]{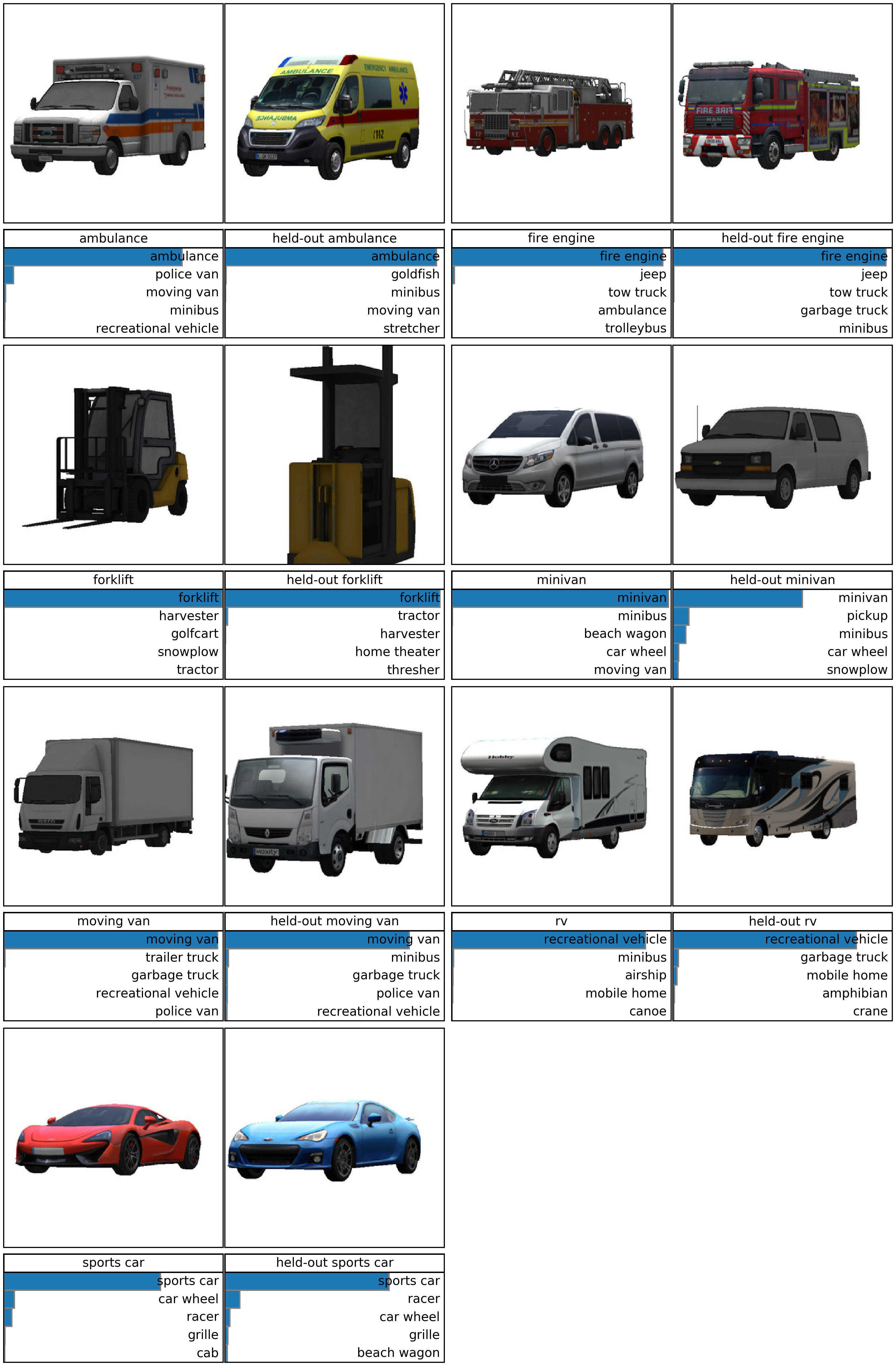}
	\caption{
	In Sec.~\ref{sec:adversarial_training}, we trained an AlexNet classifier on the 1000-class ImageNet dataset augmented with 30 additional classes that contain adversarial poses corresponding to the 30 \emph{known} objects used in the main experiments.
	We also tested this model on 7 \emph{held-out} objects.
	Here, we show the renders of 7 pairs of (training-set object, held-out object).
	The 3D objects are rendered by the NR \cite{modernGL} at a distance of $(0,0,4)$. 
	Below each image is its top-5 predictions by Inception-v3 \cite{szegedy2016rethinking}.
	The original, high-resolution figure is available at \url{https://goo.gl/Li1eKU}.}
	\label{fig:7_pairs}
\end{figure*}

\begin{figure*}
    \begin{subfigure}{\linewidth}
        \centering
        \includegraphics[width=1.0\columnwidth]{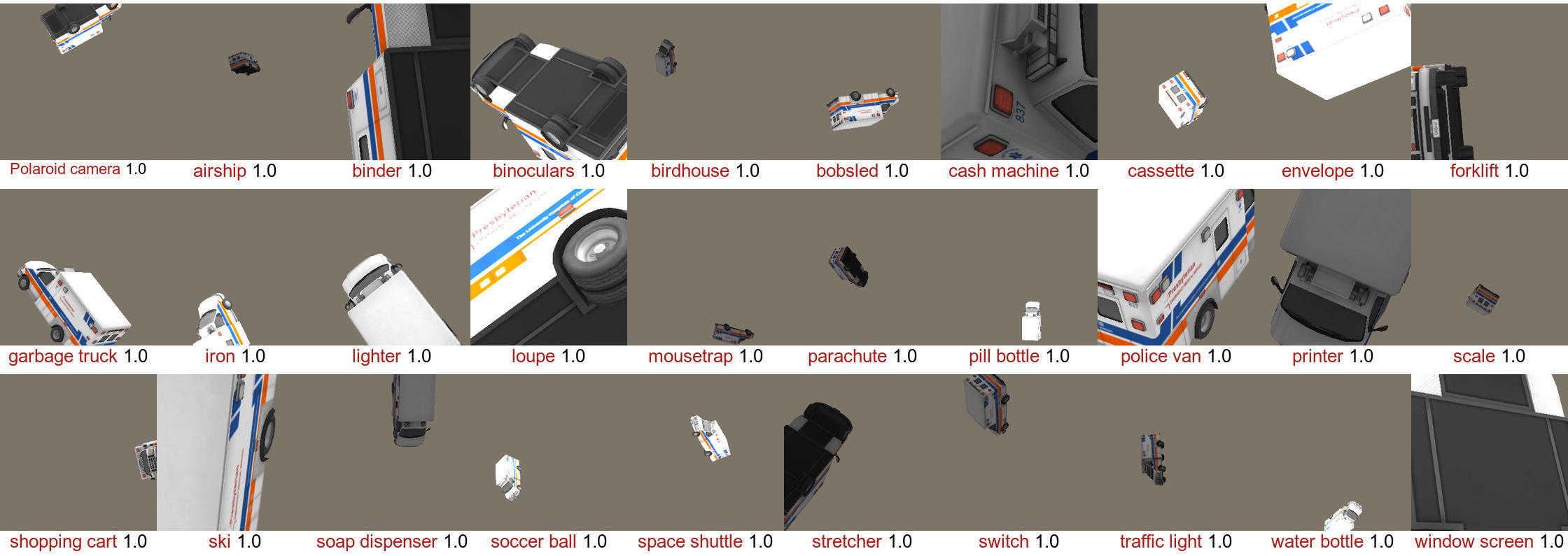}
        \caption{\class{ambulance}}\label{fig:30_ax_ambulance}
    \end{subfigure}
   \begin{subfigure}{\linewidth}
    \centering
    \includegraphics[width=1.0\columnwidth]{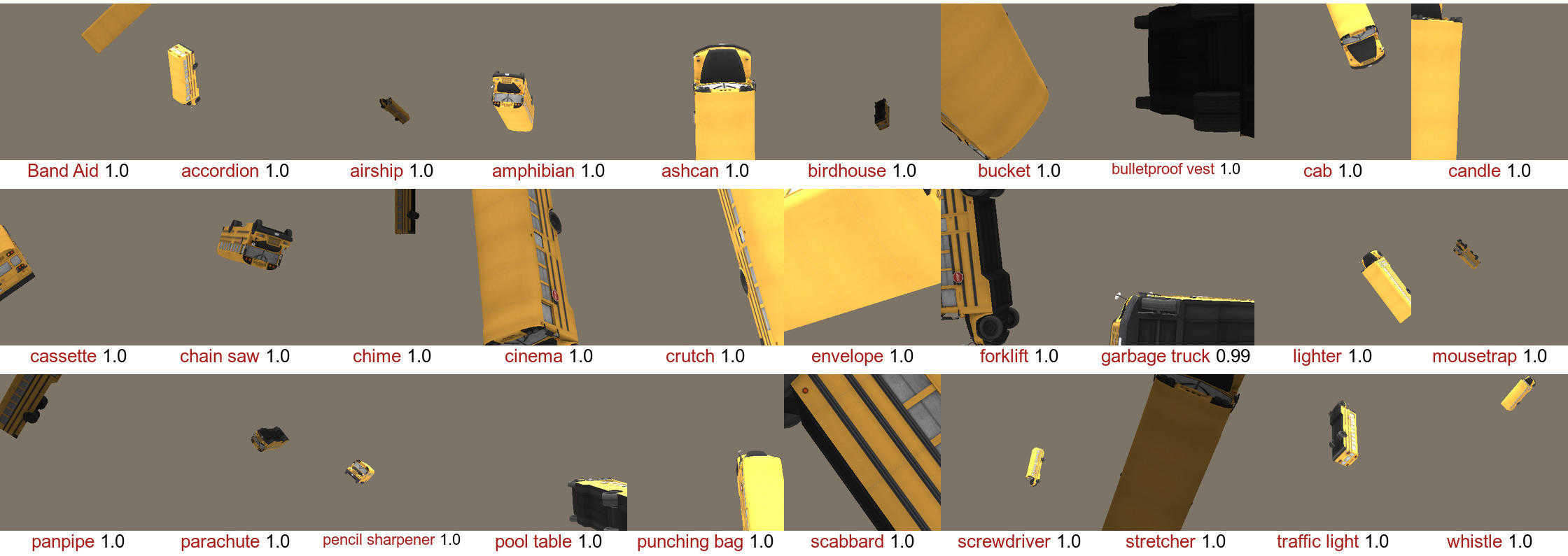}
    \caption{\class{school bus}}\label{fig:30_ax_schoolbus}
   \end{subfigure}
    \begin{subfigure}{\linewidth}
        \centering
        \includegraphics[width=1.0\columnwidth]{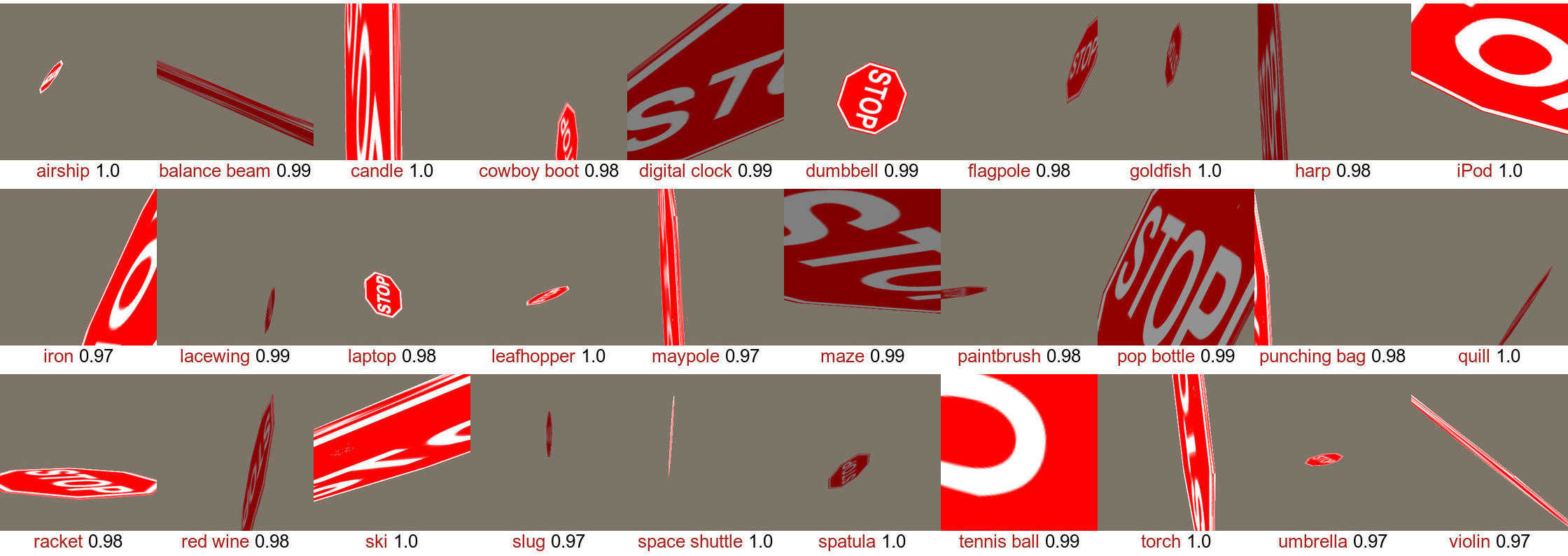}
        \caption{\class{street sign}}\label{fig:30_ax_stopsign}
    \end{subfigure}
    \caption{
    30 random adversarial examples misclassified by Inception-v3 \cite{szegedy2016rethinking} with high confidence ($p \geq 0.9$) generated from 3 objects: \class{ambulance}, \class{school bus}, and \class{street sign}. 
    Below each image is the top-1 prediction label and confidence score.
    The original, high-resolution figures for all 30 objects are available at \url{https://goo.gl/rvDzjy}.}\label{fig:30_ax}
\end{figure*}


\begin{figure*}
    \centering
    \includegraphics[width=1.0\columnwidth]{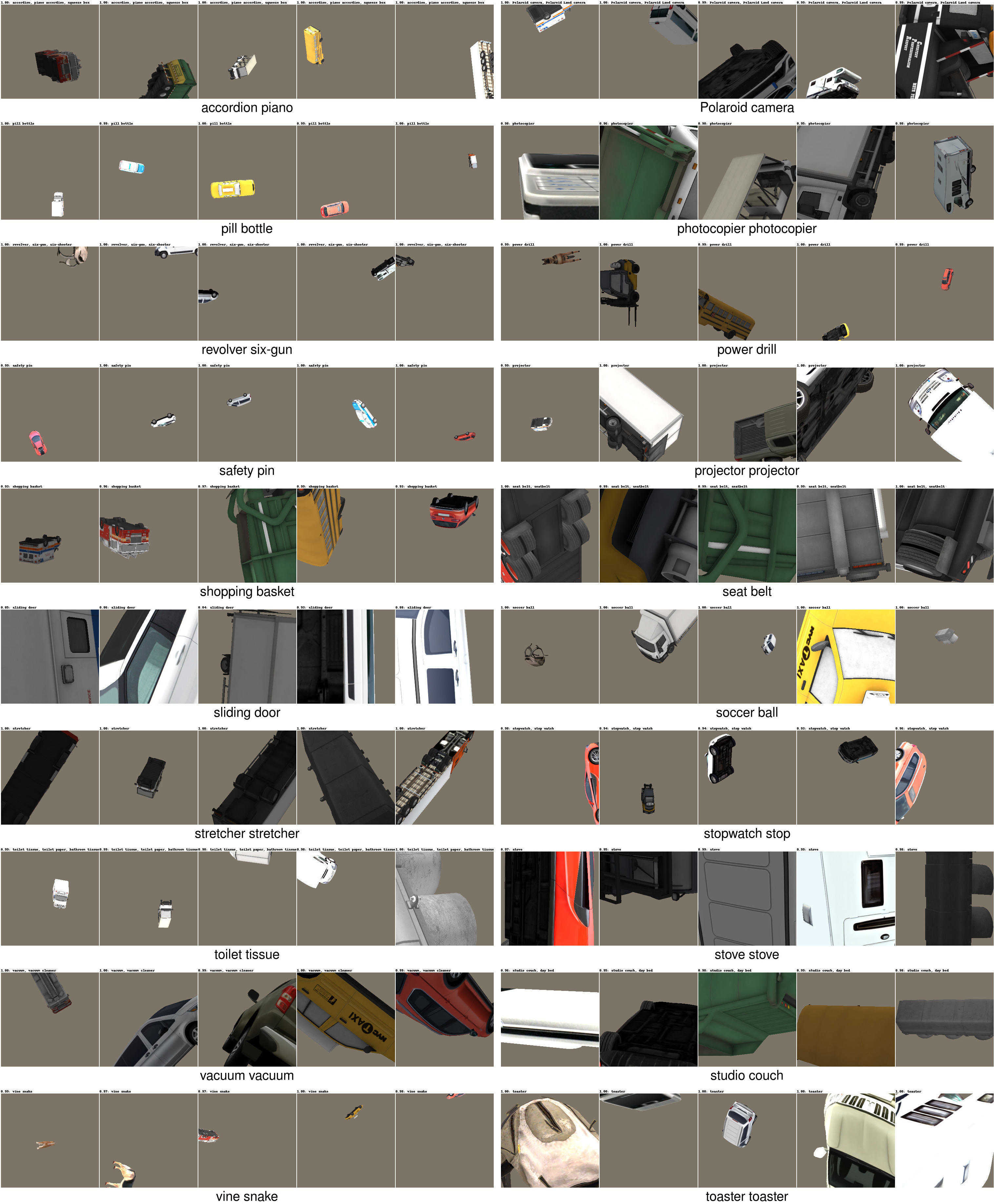}
    \caption{
    For each target class (\eg, \class{accordion piano}), we show five adversarial poses generated from five unique 3D objects.
    Adversarial poses are interestingly found to be homogeneous for some classes, \eg, \class{safety pin}.
    However, for most classes, the failure modes are heterogeneous.
	The original, high-resolution figure is available at \url{https://goo.gl/37HYcE}.
    }\label{fig:common_failures_per_label}
\end{figure*}

\newpage
\begin{figure*}[t]
    \begin{center}
        \includegraphics[width=\textwidth]{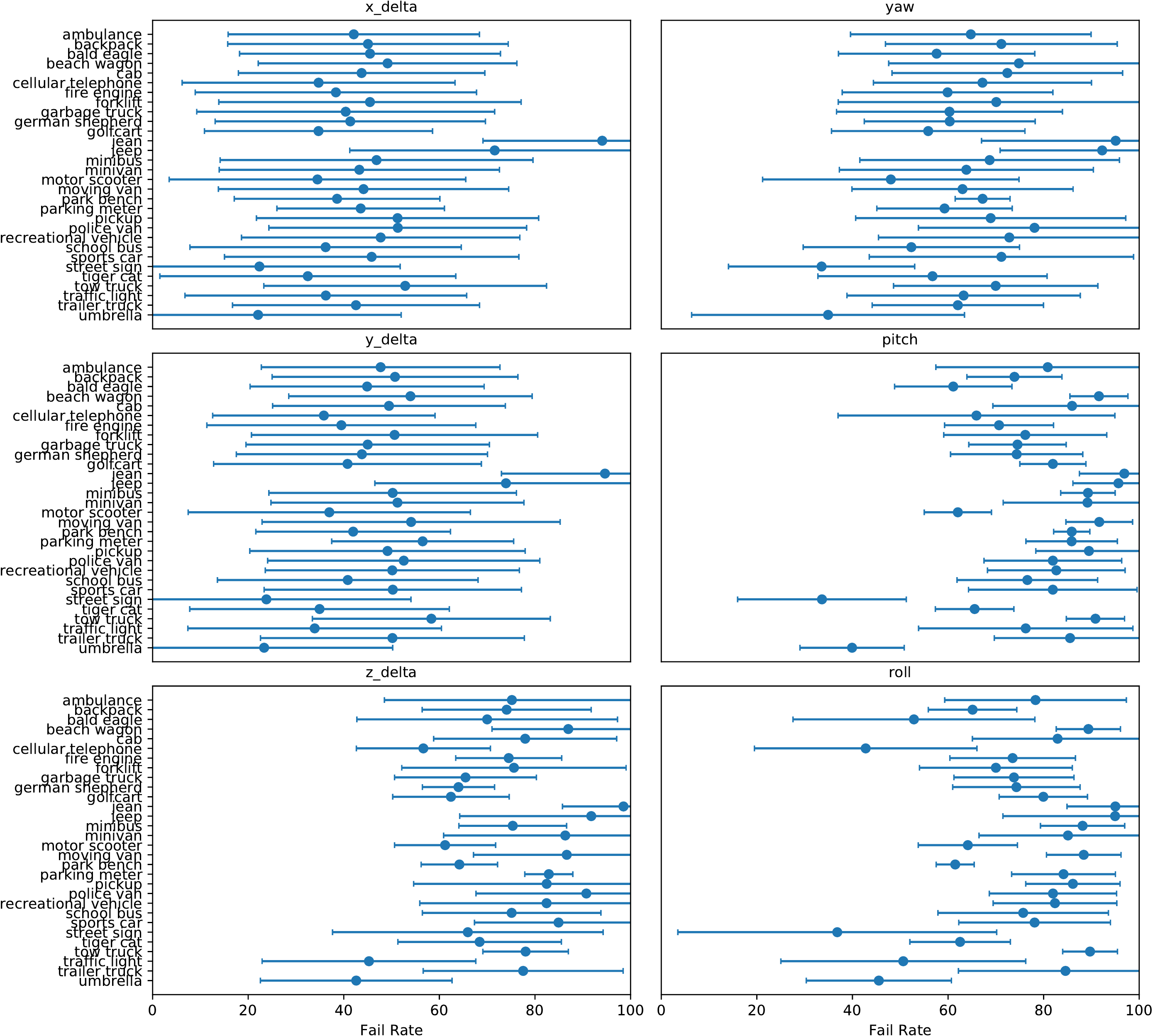}
    \end{center}
    \caption{
        Inception-v3 \cite{szegedy2016rethinking} is sensitive to single parameter disturbances of object poses that had originally been correctly classified.
        For each object, we found 100 correctly classified 6D poses via a random sampling procedure (Sec.~\ref{sec:landscape}).
        Given each such pose, we re-sampled one parameter (shown on top of each panel, \eg, yaw) 100 times, yielding 100 classifications, while holding the other five pose parameters constant.
        In each panel, for each object (\eg, \class{ambulance}), we show an error plot for all resultant $100 \times 100 = 10,000$ classifications.
        Each circle denotes the mean misclassification rate (``Fail Rate'') for each object, while the bars enclose one standard deviation.
        Across all objects, Inception-v3 is more sensitive to changes in yaw, pitch, roll, and depth (``z\_delta'') than spatial changes (``x\_delta'' and ``y\_delta'').
    }
    \label{fig:sensitivity}
\end{figure*}

\newpage

%
%

\section{DNN failure rates for ball-like objects}\label{sup:ball_objects}
To investigate the role of object geometry in DNN pose failures, we re-ran our random search procedure on five purchased ball objects: three different soccer balls, a volleyball, and a ping-pong ball.
The error rates were 4\%, 4\%, 5\%, 47\%, and 48\%, respectively; however, the incorrect labels for the volleyball and ping-pong ball objects were qualitatively often reasonable (\eg, \texttt{golf ball} and \texttt{billiard ball} for the ping-pong ball).
Clearly, the DNN can gracefully handle much of the pose space for these ``easy'' objects, but whether this robustness is due to the specific features of the classes (\eg, black and white corners where hexagons meet on a soccer ball) or data variability in the training set requires further research.

\section{Adversarial training}\label{sec:adversarial_training}

%

\subsec{Training}
We augmented the original 1,000-class ImageNet dataset with an additional 30 AX classes.
Each AX class included 1,350 randomly selected high-confidence ($p \geq 0.9$) misclassified images split 1,300/50 into training/validation sets.
Our AlexNet trained on the augmented dataset (AT) achieved a top-1 accuracy of 0.565 for the original ImageNet validation set and a top-1 accuracy\footnote{In this case, a classification was ``correct'' if it matched \emph{either} the original ImageNet positive label \emph{or} the negative, object label.} of 0.967 for the AX validation set.

\subsec{Evaluation} 
To evaluate our AT model vs. a pre-trained AlexNet (PT), we used RS to generate $10^6$ samples for each of our 3D training objects.
In addition, we collected seven held-out 3D objects not included in the training set that belong to the same classes as seven training-set objects (example renders in Fig.~\ref{fig:7_pairs}).
We followed the same sampling procedure for the held-out objects to evaluate whether our AT generalizes to unseen objects.

For each of these $30 + 7 = 37$ objects and for both the PT and our AT, we recorded two statistics: (1) the percent of misclassifications, \ie errors; and (2) the percent of high-confidence (\ie, $p \geq 0.7$) misclassifications (Table~\ref{tab:ax_stats}).

We hypothesize that augmenting the dataset with many more 3D objects 
may improve DNN generalization on held-out objects.
Here, AT might have used (1) the grey background to separate the 1,000 original ImageNet classes from the 30 AX classes; and (2) some non-geometric features 
sufficient to discriminate among only 30 objects.
However, as suggested by our work (Sec.~\ref{sec:3d_object_dataset}), acquiring a large-scale, high-quality 3D object dataset is costly and labor-intensive. 
Currently, no such public dataset exists, and thus we could not test this hypothesis.

%
%
%
%

\end{document}